\documentclass[preprint,12pt]{elsarticle}

\usepackage[pagebackref,breaklinks,colorlinks,allcolors=iccvblue]{hyperref}
\usepackage[inkscapelatex=false]{svg}
\usepackage{times}
\usepackage{latexsym}
\usepackage{booktabs}
\usepackage{pifont}
\usepackage{subcaption} 
\usepackage{svg}
\usepackage{amsmath}
\usepackage{mathtools}
\usepackage{amssymb}
\usepackage[capitalize]{cleveref} 
\usepackage[table]{xcolor}
\usepackage{mathrsfs}
\usepackage{xcolor}
\usepackage{adjustbox}
\usepackage{wrapfig}
\usepackage{algpseudocode}
\usepackage{tabularx}
\usepackage{mathbbol}
\usepackage{tcolorbox}
\usepackage{caption}
\usepackage{float}
\usepackage{lipsum}
\usepackage{xtab}
\usepackage{makecell}
\usepackage{xcolor}
\definecolor{lowcontrastgreen}{RGB}{160,206,86}
\definecolor{lowcontrastred}{RGB}{255,95,83}
\definecolor{lowcontrastyellow}{RGB}{255,218,93}
\definecolor{lowcontrastblue}{RGB}{181,234,215}
\usepackage[T1]{fontenc}
\usepackage[utf8]{inputenc}
\usepackage{boldline}
\usepackage{microtype}
\usepackage{multicol}
\usepackage{multirow}
\usepackage{inconsolata}
\usepackage{svg}
\usepackage{graphicx}
\usepackage{listings}
\usepackage{epsfig,endnotes}
\usepackage[utf8]{inputenc}
\usepackage[T1]{fontenc}
\usepackage{algorithm}
\usepackage{upquote}
\newcommand{\revise}[1]{\textcolor{black}{#1}}
\journal{Information Fusion}
\usepackage{array}
\newcolumntype{M}[1]{>{\centering\arraybackslash}m{#1}}
\usepackage{colortbl}
\usepackage{xcolor}
\definecolor{Green}{HTML}{2ECC71}
\definecolor{Red}{HTML}{E74C3C}
\newcommand{\ctick}{\textcolor{Green}{\checkmark}}
\newcommand{\ccross}{\textcolor{Red}{\ding{55}}}
\begin{document}

\begin{frontmatter}


\author[uwa,unimelb]{Yihao Ding}
\author[unimelb]{Soyeon Caren Han\corref{cor1}}
\author[unimelb,hnu]{Zechuan Li}
\author[unimelb]{Hyunsuk Chung}

\cortext[cor1]{Corresponding author: caren.han@unimelb.edu.au}


\affiliation[uwa]{
  organization={The University of Western Australia}, 
  city={Perth},
  state={WA},
  country={Australia}
}
\affiliation[unimelb]{
  organization={The University of Melbourne}, 
  city={Melbourne},
  state={VIC},
  country={Australia}
}
\affiliation[hnu]{
  organization={Hunan University},
  city={Changsha},
  country={China}
}

\title{SynJAC: Synthetic-data-driven Joint-granular Adaptation and Calibration for Domain Specific Scanned Document Key Information Extraction}




\begin{abstract}
Visually Rich Documents (VRDs), comprising elements such as charts, tables, and paragraphs, convey complex information across diverse domains. However, extracting key information from these documents remains labour-intensive, particularly for scanned formats with inconsistent layouts and domain-specific requirements. Despite advances in pretrained models for VRD understanding, their dependence on large annotated datasets for fine-tuning hinders scalability. This paper proposes \textbf{SynJAC} (Synthetic-data-driven Joint-granular Adaptation and Calibration), a method for key information extraction in scanned documents. SynJAC leverages synthetic, machine-generated data for domain adaptation and employs calibration on a small, manually annotated dataset to mitigate noise. By integrating fine-grained and coarse-grained document representation learning, SynJAC significantly reduces the need for extensive manual labelling while achieving competitive performance. Extensive experiments demonstrate its effectiveness in domain-specific and scanned VRD scenarios.
\end{abstract}






\end{frontmatter}



\section{Introduction}
\label{sec:intro}

Visually Rich Documents (VRDs) containing numerically qualified and potentially sensitive information are typically shared intra-departmentally or between institutions rather than being publicly accessible. Automatically extracting information precisely and economically from domain knowledge-intensive documents is challenging, especially given the rapidly increasing demands across multiple domains such as finance \citep{gerling2025multimodal} and education \citep{vies}, unlike highly qualified academic papers \citep{mmvqa}, the flexible formats of scanned documents further complicate the task. To meet these demands, various pretrained VRD understanding frameworks \citep{layoutlmv3, lilt} leverage self-supervised pretraining to capture general document domain knowledge. However, deploying these frameworks effectively in real-world scenarios often requires extensive domain-specific annotations from experts, which can be labour-intensive and time-consuming, potentially delaying projects and hindering practical deployment.

\begin{figure}[t]
  \centering
   \includegraphics[width=\textwidth]{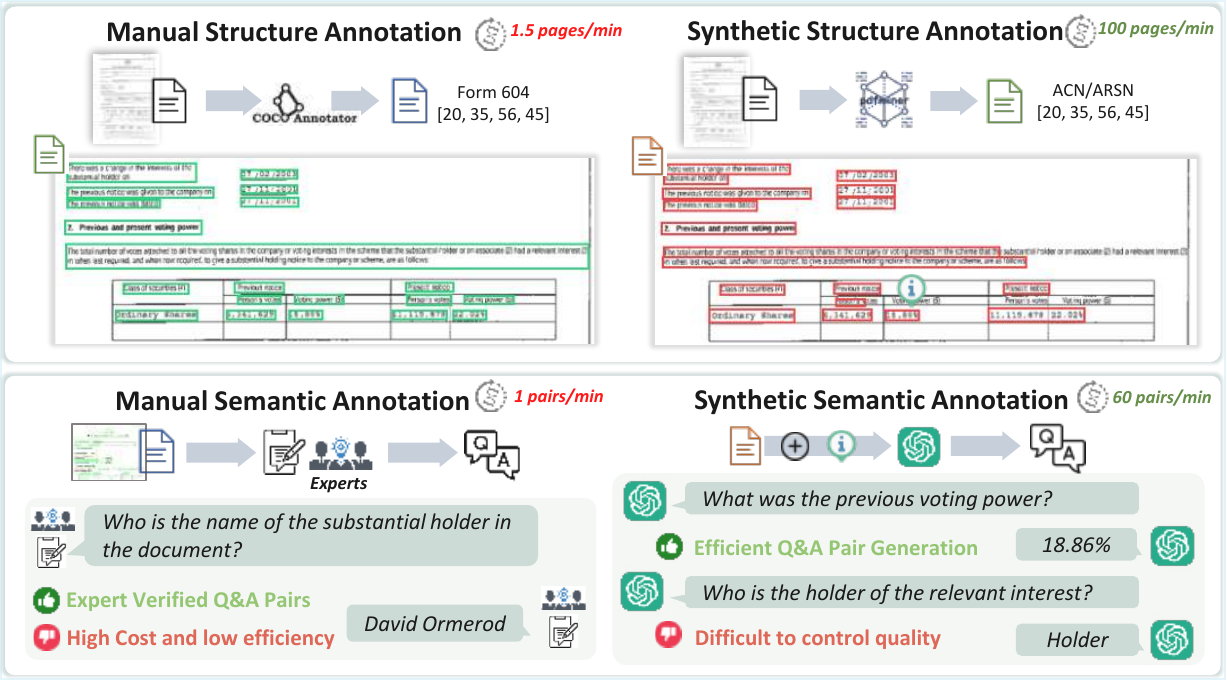}
  \caption{Comparing manual and synthetic structural and task-oriented annotation. }
  \label{fig:introduction}
  \vspace{-0.5cm}
\end{figure}

From a human perspective, understanding a document in a new domain typically begins with analysing its layout structure, followed by content interpretation based on user intent. Substantial manual annotation is often necessary to capture domain-specific layout structures \citep{doclaynet}, which in turn supports layout analysis models \citep{pena2024continuous,zhudeformable}. However, producing high-quality annotations with detailed structural and semantic information (as illustrated in Figure~\ref{fig:introduction}) is both time-consuming and labour-intensive, particularly for scanned documents containing noise, handwritten content, and inconsistent formatting. While PDF parsers and OCR systems can generate large-scale, coarse structural and semantic annotations efficiently, their outputs are often noisy and imprecise. Despite this, the potential of synthetically generated structure annotations remains under-explored, especially in the context of scanned, unstructured VRDs. Synthetic annotations can offer rich contextual representations of document structure and layout, tailored to the target domain, thereby facilitating more robust domain adaptation and mitigating the dependency on manual labelling.


An adequate understanding of document content often requires training models on task-specific and well-annotated datasets tailored to end-user needs. Various manually annotated datasets have been designed for key information extraction (KIE) across domains like finance \citep{formnlu}, education \citep{vies}, and scanned receipts \citep{cord}. Creating these annotations often demands domain expertise to align content with user requirements and typically involves preliminary layout annotation, as shown in Figure~\ref{fig:introduction}. However, real-world VRDU solutions should reduce reliance on labour-intensive annotations by enabling deep learning frameworks to achieve competitive performance with minimal manual effort. Large language models (LLMs) \citep{chatgpt} and multimodal large language models (MLLMs) \citep{qwen} have shown significant advancements in zero-shot VRDU tasks \cite{docvqa} and facilitate VRD question answering dataset generation via prompt engineering \citep{mmvqa}, leveraging extensive training on diverse corpora. Nevertheless, the potential of using synthetic content annotation to tackle domain-specific VRD in real-world KIE applications remains largely under-explored.

\revise{This paper presents SynJAC, a Synthetic-data-driven Joint-granular Adaptation and Calibration framework that achieves competitive performance on KIE tasks with only a limited number of manually annotated documents, especially on noise scanned documents. SynJAC begins by generating synthetic structural, semantic, and task-oriented annotations using off-the-shelf tools and LLMs.
To effectively distil knowledge from the generated synthetic data, we introduce a Joint-grained Model that leverages pretrained backbones to encode both fine-grained (word-level) and coarse-grained (entity-level) features, thereby capturing implicit general domain knowledge. A novel layout encoding method, Layout-to-Vector (L2V), is further proposed to enhance spatial layout representation and strengthen inter-grained correlations.
To mitigate domain shift, we introduce a suite of \textbf{Domain Adaptation Strategies} that adaptively train the joint-grained model on synthetic data, achieving both structural alignment (\textit{Structural Domain Shifting}) and task-specific adaptation (\textit{Synthetic Sequence Tagging} and \textit{Synthetic Instruction Tuning}) toward the target document collection. Finally, a \textbf{Domain Calibration} module refines this adaptation by leveraging a small set of high-quality annotations and pooling mechanisms to effectively balance curated and synthetic knowledge.}

\revise{The main contributions of this work are summarized as follows:
 (1) We propose \textbf{SynJAC}, a unified framework designed to enable robust key information extraction in domain-specific scanned VRDs with minimal manual supervision.
 (2) SynJAC incorporates a synthetic annotation pipeline that leverages off-the-shelf tools and LLMs to generate structural and task-specific annotations, significantly reducing manual annotation costs.
 (3) A joint-grained architecture integrated with a new layout embedding method (L2V) is proposed, which captures comprehensive document representations by enhancing both inter- and intra-grained feature learning.
 (4) A suite of domain adaptation strategies is designed to mitigate domain gaps and enhance inter-grained feature representations including structural alignment and task specific adaptation.
 (5) Extensive experiments on unstructured and scanned documents demonstrate the effectiveness of SynJAC across diverse KIE scenarios.}

\section{Related Work}
\label{sec:related_work}

\begin{table}[t]
\centering
\begin{adjustbox}{max width=\linewidth}
\begin{tabular}{llcccc}
\toprule
\textbf{Type} & \textbf{Model} & \textbf{General} & \textbf{Syn-Str.} & \textbf{Syn-Sem.} & \textbf{Manual} \\
\midrule

\multirow{1}{*}{Heuristic}
  & Watanabe et al. \cite{watanabe1995layout} & \ccross & \ccross & \ccross & Large \\
\midrule
\multirow{2}{*}{Feature Driven}
  & VIES       \cite{vies}     & \ccross & \ccross & \ccross & Large \\
  & Chen et al.  \cite{chen2023task}   & \ccross & \ccross & \ccross & Few/Zero-shot \\
\midrule
\multirow{2}{*}{\centering Fine-grained}
  & LayoutLMv3  \cite{layoutlmv3}    & \ctick  & \ccross & \ccross & Large \\
  & LiLT        \cite{lilt}    & \ctick  & \ccross & \ccross & Large \\
\midrule
\multirow{2}{*}{Coarse-grained}
  & SelfDoc  \cite{selfdoc}       & \ctick  & \ccross & \ccross & Large \\
  & UniDoc   \cite{udoc}       & \ctick  & \ccross & \ccross & Large \\
\midrule
\multirow{1}{*}{Joint-grained}
  & 3MVRD     \cite{m3vrd}      & \ctick  & \ccross & \ccross & Large \\
\midrule
\multirow{2}{*}{MLLMs}
  & LayoutLLM   \cite{luo2024layoutllm}    & \ctick  & \ccross & \ccross & Few/Zero-shot \\
  & Docrouter  \cite{zhang2025docrouter}  & \ctick  & \ccross & \ccross & Few/Zero-shot \\
\midrule
\multirow{2}{*}{Synthetic Driven}
  & DocKD   \cite{kim2024dockd}        & \ctick  & \ccross & \ctick  & Zero-shot \\
  & SynJAC (Ours)         & \ctick  & \ctick  & \ctick  & Few/Zero-shot \\

\bottomrule
\end{tabular}
\end{adjustbox}
\caption{\revise{Comparison of knowledge integration across VRDU models. Each model is evaluated based on its use of general knowledge, synthetic structure-aware (Syn-Str.) and semantic (Syn-Sem.) knowledge, and reliance on manual annotations. "Large" indicates full supervision is required, while "Few/Zero-shot" denotes minimal or no manual labelling.}}
\label{tab:related_work}
\end{table}
\paragraph{\textbf{Visually-Rich Document Understanding}} Heuristic methods \citep{watanabe1995layout} and statistical machine learning \citep{oliveira2017fast} were applied to closed-domain document applications but required expert customization. Feature-driven approaches \citep{chen2023task, vies} leverage multimodal feature representations to enhance generalizability but lack general-knowledge integration, limiting their performance potential. Recent advances in layout-aware pre-trained frameworks \citep{layoutlmv3, lilt} and joint-grained models \citep{lyu2024structextv3} demonstrate improved document representation through large-scale general-domain self-supervised pretraining, yet they depend heavily on extensive annotated data for effective domain-specific knowledge transfer. LLM/MLLM-based frameworks \citep{zhang2025docrouter, luo2024layoutllm} have demonstrated improved zero/few-shot performance for VRD understanding tasks by leveraging broad pretraining and instruction-tuning on general domain. However, the reliance on large annotated datasets remains a barrier, as shown by Table~\ref{tab:related_work}, underscoring the need for scalable solutions like synthetic data generation, as explored in this paper.

\paragraph{\textbf{Domain Adaptation and Knowledge Distillation}} Domain adaptation is crucial in transfer learning, encompassing several variants such as unsupervised domain adaptation \citep{wang2020transfer} and source-free domain adaptation \citep{liang2020we}, which focus on transferring knowledge from one source domain to a target domain that differs from our scenarios. Another subproblem within transfer learning, knowledge distillation\citep{hinton2015distilling}, involves transferring knowledge from a large-scale teacher to small student networks. This has been widely applied in language \citep{adhikari2020exploring}, vision \citep{fang2021compressing}, and multimodal applications \citep{ma2023using}, yet there is a lack of research exploring knowledge distillation in VRDU. While some efforts \citep{m3vrd} have explored joint-grained knowledge distillation for VRDU, they continue to rely heavily on large, annotated datasets and require extensive fine-tuning for practical use. \revise{Another recent study \cite{kim2024dockd} explores synthetic question-answer pairs to integrate domain-specific knowledge but overlooks structural information in mitigating the domain gap}. Our work addresses this limitation by leveraging synthetic data for domain adaptation from both structural and semantic perspectives, complemented by calibration with minimal manual annotation, achieving competitive results without reliance on large-scale labelled data.
\section{Problem Formulation} 
\label{sec:problem_formulation}
\paragraph{\textbf{Preliminary Definition}} Information in a visually rich document (VRD) can be organized and modelled at both fine-grained and coarse-grained levels \cite{udoc,m3vrd}. The fine-grained level focuses on individual text tokens along with their content and spatial layout, while the coarse-grained level captures higher-level semantic structures, such as paragraphs, tables, and figures, by aggregating textual, layout, and visual cues. The task of Key Information Extraction (KIE) can operate at either granularity \cite{ding2024deep}: at the fine-grained level, it could be sequence tagging task to classify each token into a predefined category; at the coarse-grained level, it involves identifying the most relevant semantic entity for a given key. 

\paragraph{\textbf{Task Formulation}} This paper addresses the practical and challenging task of extracting predefined key information from a collection of domain-specific documents $\mathbb{D}$, under the constraint of limited manual annotations. To tackle this problem, we partition the document collection into three subsets: a large unannotated set $\mathbb{D}_s$, and a small manually annotated subset, which is further divided into a guidance set $\mathbb{D}_g$ and a test set $\mathbb{D}_t$.
The unannotated set $\mathbb{D}_s$ is leveraged to generate synthetic training data, which is used to adapt pretrained models and reduce the domain gap between general-purpose pretraining corpora and the target domain. To mitigate the impact of noise introduced by the synthetic data, the guidance set $\mathbb{D}_g$ is employed to further refine the model. The test set $\mathbb{D}_t$ is used for reporting the performance in this paper.

\section{Methodology}
This section introduces the \textbf{SynJAC}(Synthetic-data-driven Joint-granular Adaptation
and Calibration) framework, which consists of four key components: \textit{Synthetic Data Generation}, \textit{Joint-grained Model Architecture}, \textit{Domain Adaptation Strategies}, and \textit{Guidance-Based Domain Calibration}.
During the synthetic data generation stage, the unannotated dataset $\mathbb{D}_s$ is processed using off-the-shelf OCR tools to extract layout information and LLMs to generate task-oriented annotations at both fine-grained and coarse-grained levels.
With these synthetic annotations, $\mathbb{D}_s$ is then used to train a joint-grained model $\mathcal{F}$, which captures and encodes inter/intra grained representations.
To improve generalization to the target domain, we introduce diverse domain adaptation strategies: \textit{Structural Domain Shifting} (SDS) aligns document structures across domains, while \textit{Synthetic Sequence Tagging} (SST) and \textit{Synthetic Instruction-Tuning} (SIT) enable task-specific adaptation for fine- and coarse-grained KIE tasks, respectively.
Finally, in the guidance-based adaptation, the tuned model $\mathcal{F}$ is further refined using the guidance set, a small, manually annotated subset, helping to mitigate noise in the synthetic annotations and enhance overall performance.
The following sections describe each stage of the framework in detail.

\subsection{Synthetic Data Generation}
\begin{figure*}[t]
  \centering
  \includegraphics[width=\linewidth]{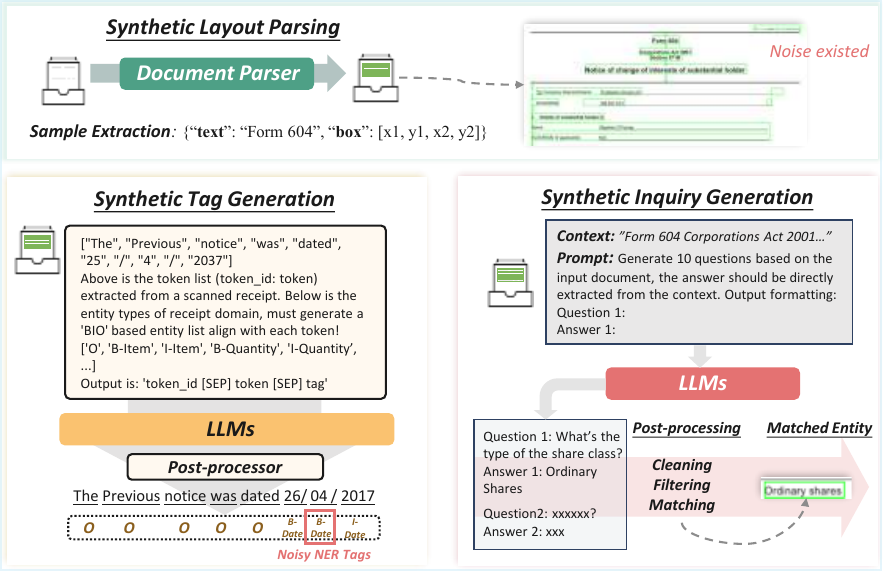}
  \caption{Workflow for generating synthetic annotations for domain-specific understanding.}
  \label{fig:data_prepare}

\end{figure*}

Acquiring high-quality manually annotated data is labour-intensive and costly, as illustrated in Figure~\ref{fig:introduction}. Effective adaptation of the framework to target document collections requires fine-tuning on supervised or self-supervised tasks specific to the domain, supported by an adequate dataset (see Section~\ref{sec:domain_adaptation}). To reduce the manual effort required for domain adaptation, we introduce a synthetic data generation workflow that produces both structural and task-specific annotations for document collections, as shown in Figure~\ref{fig:data_prepare}. After obtaining a specific document collection, we divide it into three subsets as previously described: a large-scale unannotated set $\mathbb{D}_s$, a manually annotated guidance set $\mathbb{D}_g$ for adaptation, and a test set $\mathbb{D}_t$ for evaluation. Next, we apply synthetic layout parsing to derive layout annotations on unannotated set $\mathbb{D}_s$, followed by synthetic tag generation and synthetic inquiry generation, which produce task-oriented labels for adapting fine-grained and coarse-grained VRD-KIE, respectively.

\paragraph{\textbf{Synthetic Layout Parsing}} It aims to automatically extract the layout structure of documents by identifying text lines and semantic entities, while extracting their associated textual content. The document image is processed using off-the-shelf parsers, such as PDFMiner or OCR tools\footnote{For example, PaddleOCR: \url{https://github.com/PaddlePaddle/PaddleOCR}.}, to automatically produce layout annotations. These annotations typically include bounding box coordinates and the corresponding text, as illustrated in Figure~\ref{fig:data_prepare}. 
    
\paragraph{\textbf{Synthetic Tag Generation}} This module aims to produce synthetic labels for word sequences to support sequence tagging-based KIE tasks. A LLM, such as ChatGPT~\cite{chatgpt}, takes as input a list of text tokens and a predefined label set (e.g., BIO tags) of a target document, and is prompted to assign a tag to each token. The model is required to output each token's index, content, and predicted tag, separated by “[SEP]”, to facilitate accurate post-processing. For example, given the input:
    \texttt{["26", "/", "04", "/", "2017"]},
    the post-processed synthetically generated labels: 
    \texttt{[B-Date, I-Date, B-Date, I-Date, I-Date]}. Although these synthetic tags may be noisy, adaptively tuning the model on this data can enhance domain adaptation and improve fine-grained contextual understanding for scanned document KIE.

\paragraph{\textbf{Synthetic Inquiry Generation}} We leverage LLMs to automatically generate QA pairs from document content, following established practices in VRD-QA dataset construction~\cite{mmvqa}. For each document image, the LLM is prompted with its extracted textual content and instructed to generate multiple questions, with the constraint that answers must be directly retrievable from the context.
After generation, a post-processing pipeline is applied to clean, filter, and validate the QA pairs. Each generated answer is then semantically aligned with the set of layout-parsed semantic entities from the document using fuzzy string matching. For example, given a generated QA pair:
\texttt{Question: What’s the type of the share class?}
\texttt{Answer: Ordinary Shares}, the answer is closest fuzzy matched semantic entity, “Ordinary shares” (as represented in Figure~\ref{fig:data_prepare}), which is then designated as the retrieval target during domain adaptation.

\subsection{Joint-grained Model Architecture}
As introduced in Section~\ref{sec:problem_formulation}, documents can be modelled at both fine-grained and coarse-grained levels. Fine-grained representations capture detailed textual and structural features, while coarse-grained representations are better suited for modelling layout and logical relationships. To obtain a comprehensive document representation, we design a joint-grained architecture $\mathcal{F}$ that facilitates inter- and intra-granular interactive learning. The initial document representation is first encoded by pretrained fine- and coarse-grained models to extract multimodal features. A joint-grained encoder $\mathcal{E}_{jg}$ then enhances inter-granularity and structure-aware understanding. Transformer decoders of both fine-grained $\mathcal{E}_{fg}$ and coarse-grained $\mathcal{E}_{cg}$ refine these features into task-specific representations, and linear or transformer based adaptors are supporting domain calibration by tuning on a small guidance set to reduce noise from synthetic data.
\begin{figure*}[t]
  \centering
  \includegraphics[width=0.9\linewidth]{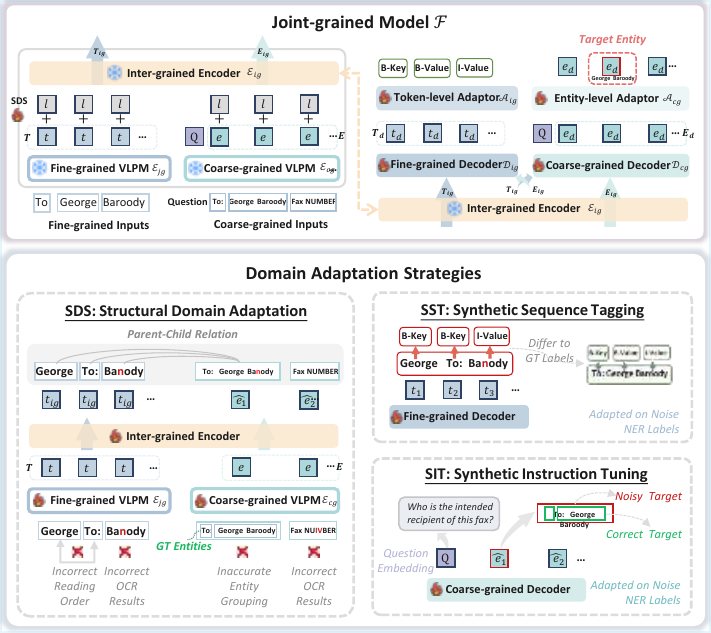}
  \caption{The SynJAC framework is built on a Joint-grained Model containing both fine-grained and coarse-grained document representations (left). We introduce three domain adaptation strategies, SDS, SST, and SIT, to enable the joint-grained framework to effectively adapt to the target domain from both structural and task-oriented perspectives.}
  \label{fig:domain_framework}
\end{figure*}
\paragraph{\textbf{Initial Document Representation}}
Before introducing the joint-granular model architecture, $\mathcal{F}$, we first describe the feature encoding methods used to represent multimodal information. Each manually or synthetically annotated document is represented as a sequence of textual tokens (fine-grained), along with a set of semantic entities or text lines (coarse-grained), where each unit contains textual content and corresponding bounding box coordinates. To extract fine-grained token-level features, we employ a layout-aware Vision-Language Pre-Trained model (VLPM) $\mathcal{E}_{fg}$ (e.g., LayoutLMv3~\cite{layoutlmv3}, LiLT~\cite{lilt}), which encodes the multimodal input including text (both question and context), bounding box, and document image, to produce token representations $T = \{t_1, \dots, t_n\}$. For each semantic entity, we follow the approach in \cite{docgcn} to obtain a semantic embedding $s_i$ using a language encoder (e.g BERT \cite{bert}) and a visual embedding $v_i$ using a vision backbone (e.g ResNet-50 \cite{fasterrcnn}. The coarse-grained VLPM $\mathcal{E}_{cg}$ (e.g. LXMERT \cite{lxmert}, VisualBERT \cite{visualbert}) is applied to get encoded query text sequence $Q$ (e.g. "Company Name") and the enhanced visual feature. The final entity representation is obtained by concatenating the enhanced visual and semantic embeddings: $E = \{e_1, \dots, e_m\}$, where $e_i = [\mathcal{E}_{cg}(v_i); s_i]$.

 To further incorporate layout information to boost domain adaptation and enhance the modelling of inter-grained relationships, we propose \textbf{Layout-to-Vector} (L2V), a novel method that transforms spatial layout cues into visual signals by rendering each document into a colour-coded image $\mathbf{I}_{layout}$ based on the $x$ and $y$ coordinates of its elements. For each pixel located at position $(x_j, y_k)$ in an image of size $W \times H$, we have a pseudo-RGB vector:
\begin{equation}
    \mathbf{p} = \left[255 \cdot \tfrac{x_j}{W},\; 255 \cdot \tfrac{y_k}{H},\; 255\right],
\end{equation}

where the red and green channels correspond to horizontal and vertical spatial cues, respectively. To obtain the layout embedding $l_i$ for $i$-th token or entity object, we apply RoIAlign over the pretrained CNN feature map of $\mathbf{I}_{layout}$ at the region corresponding to the object's bounding box $b_i$:
\begin{equation}
l_i = \text{RoIAlign}\left( \text{CNN}(\mathbf{I}_{\text{layout}}), b_i \right), \quad \mathbf{I}_{\text{layout}} \in \mathbb{R}^{H \times W \times 3},
\end{equation}
where $\mathbf{I}_{\text{layout}}$ is the colour-coded layout image and $\text{bbox}_i$ is the region corresponding to the $i$-th token or entity. This enables the model to explicitly encode spatial dependencies and visual regularities as visual features through convolutional patterns.

\paragraph{\textbf{Inter-grained Encoder}} We adopt an inter-grained encoder $\mathcal{E}_{\text{ig}}$ to enable iterative interaction and spatial reasoning between fine-grained token-level and coarse-grained entity-level representations (e.g., semantic entities or text lines). Specifically, the fine-grained token representation $T$ and coarse-grained entity representation $E$, both enhanced by VLPMs, are further fused with their respective layout-to-visual (L2V) features, denoted as $L_{\text{fg}}$ and $L_{\text{cg}}$. These enriched features are fed into the inter-grained encoder to produce refined representations:
\begin{equation}
[T_{\text{ig}},\; E_{\text{ig}}] = \mathcal{E}_{\text{ig}}([T + L_{\text{fg}};\; E + L_{\text{cg}}]),
\end{equation}
where $T_{\text{ig}}$ and $E_{\text{ig}}$ denote the inter-grained representations for tokens and entities, respectively.

\paragraph{\textbf{Task-oriented Modules}} To enable the joing-grained model 
To boost iterative inter- and intra-granular learning, enabling better adaptation of the enhanced representations to downstream tasks, the fine-grained $\mathcal{D}_{fg}$ and coarse-grained $\mathcal{D}_{cg}$ decoders take their respective granularity-level representations as inputs and treat the counterpart representations as contextual sources. Then, we could get enhanced fine-grained ($T_{d}$) and coarse-grained ($E_{d}$) representations: 
\begin{equation}
T_{d} = \mathcal{D}_{fg}(T_{ig} \parallel E_{ig}), \quad 
E_{d} = \mathcal{D}_{cg}(E_{ig} \parallel [Q; T_{ig}])
\end{equation}

The enhanced features are adaptively enhanced on task-oriented domain adaptation tasks to improve their suitability for downstream tasks. Moreover, to mitigate potential noise introduced by synthetically generated data, we incorporate MLP-based adaptors, $\mathcal{A}_{fg}$ and $\mathcal{A}_{cg}$, which are only applied during tuning on the guidance set $\mathbb{D}_g$. The detailed task-oriented domain adaptation and guidance-based calibration procedures are described in Section~\ref{sec:domain_adaptation} and Section~\ref{sec:calibration}.

\subsection{Domain Adaptation Strategies} 
\label{sec:domain_adaptation}
To facilitate the framework's comprehension of structural, semantic, and task-specific characteristics within the target domain, we employ domain adaptation strategies. These allow the joint-grained framework $\mathcal{F}$ to adapt to target document collections using synthetically annotated data $\mathbb{D}_s$, thereby obviating the need for extensive manual annotation. A Structural Domain Shifting (SDS) method captures structural and semantic patterns while enhancing inter-grained learning. Additionally, two task-specific adaptation techniques, Synthetic Sequence Tagging (SST) and Synthetic Instruction Tuning (SIT), further promote interactive learning between fine, and coarse-grained features, facilitating adaptation to downstream KIE tasks. The proposed tasks effectively bridge domain gaps between general corpora and target document collections by aligning structural, semantic, and task-specific features.

\paragraph{\textbf{Structural Domain Shifting (SDS)}} 
To effectively perform content understanding tasks like KIE or QA, it is crucial to capture both structural and semantic information from target domain documents. Thus, we introduce Structural Domain Shifting (SDS), to predict the existence of parent-child relationships between paired fine-grained tokens and coarse-grained entities. To boost the inter-grained interactively learning, SDS is performed on the outputs from inter-grained encoder, $\mathcal{E}_{ig}$.
If a parent-child relation $r$ exists between token $t_i \in T_{ig}$ and entity $e_j \in E_{ig}$, we set $r_{t_i, e_j} = 1$; otherwise, $r_{t_i, e_j} = 0$, where $T_{ig}$ and $E_{ig}$ are the output representations from the encoder $\mathcal{E}_{ig}$. The predicted relation score is computed as the dot product:$\gamma_{t_i, e_j} = \text{Linear}(t_i) \cdot \text{Linear}(e_j),$
where $\cdot$ denotes the dot product.
All scores across the $n$ tokens and $m$ entities are aggregated into a ground truth relation matrix $\hat{M}_{t,e} \in \mathbb{R}^{n \times m}$ and compared with the predicted matrix $M_{t,e} \in \mathbb{R}^{n \times m}$. The training objective of SDS is to minimize the mean squared error between these two matrices:
\begin{equation}
\arg \min_\theta \, \mathcal{L}_{\text{MSE}}\left(p(M_{t,e} \mid \theta),\, p(M'_{t,e})\right).
\end{equation}

By enforcing alignment between fine-grained token features and their corresponding coarse-grained entity representations, SDS guides the model to capture structural dependencies, while the entity-level context enriches semantic understanding based on the associated fine-grained text span.

\paragraph{\textbf{Synthetic Sequence Tagging (SST)}}
SST is introduced to enable multimodal fine-grained features to adapt to the sequence tagging style required for KIE in target domain document collections. It operates on the $n$ fine-grained textual token features $T_d = \{t_1, \dots, t_n\}$ produced by the decoder $\mathcal{D}_{fg}$, which are then fed into a linear layer to predict the logits $\hat{Y}_{fg}$: $\hat{Y}_{fg} = \text{Linear}(T_d).$
Supposing $Y_{fg} = \{y_1, \dots, y_n\}$ is the set of synthetic sequence labels for the corresponding tokens, the training objective is to minimize the cross-entropy (CE) loss between the predicted logits $\hat{Y}_{fg}$ and the synthetic labels $Y_{fg}$:
\begin{equation}
\arg\min_{\theta} \, \mathcal{L}_{\text{CE}}(p(\hat{Y}_{fg} \mid T_d),\, p(Y_{fg})).
\end{equation}

Performing SST leverages task-oriented synthetic annotations (i.e., pseudo tags) to effectively adapt the framework to fine-grained KIE tasks, while also enhancing task-specific inter-grained feature fusion.

\paragraph{\textbf{Synthetic Instruction Tuning (SIT)}} 
SIT is introduced to adapt the framework for coarse-grained KIE tasks. It operates on the coarse-grained entity features $E_d = \{e_1, \dots, e_m\}$ output by the coarse-grained decoder $\mathcal{D}_{cg}$. Given a query $Q$, the goal is to identify the target entity $e_i \in E_d$ that best matches the answer. The ground truth label $Y_{cg}$ is represented as a one-hot vector, where $y_{cg_i} = 1$ indicates the correct entity. 
The prediction is made by first applying a linear projection to each entity representation, followed by a Pointer Network (PN) \cite{formnlu} to compute the selection logits:
$
\hat{Y}_{cg} = \text{PN}(\text{Linear}(E_d)).
$
The training objective is to minimize the cross-entropy (CE) loss between the predicted logits $\hat{Y}_{cg}$ and the ground truth labels $Y_{cg}$:
\begin{equation}
\arg\min_\theta \, \mathcal{L}_{\text{CE}}(p(\hat{Y}_{cg} \mid E_d),\, p(Y_{cg})).
\end{equation}

\subsection{Guidance-based Domain Calibration}
\label{sec:calibration}
To adapt the framework from general-domain pretraining to a specific target domain, we employ the domain adaptation strategies described in Section~\ref{sec:domain_adaptation}. Although synthetic data effectively simulate task-specific scenarios, they often introduce structural and semantic noise, as illustrated in Figure~\ref{fig:introduction}. To address this, we incorporate a small, manually annotated guidance set $\mathbb{D}_g$ to calibrate the model through further fine-tuning. This calibration is applied to both coarse-grained and fine-grained adaptors, thereby refining key information extraction (KIE) accuracy at different granularity levels. Additionally, to prevent overfitting during calibration, we employ a pooling strategy that selects the most informative representations across different stages, balancing synthetic domain knowledge with the carefully curated guidance set.

For fine-grained domain calibration, the final prediction $\tilde{Y}_{fg}$ is obtained by applying max-pooling to fine-grained encoder outputs ($T$), inter-grained encoder outputs ($T_{ig}$), and fine-grained decoder outputs ($T_d$), followed by a linear adaptor $\mathcal{A}_{fg}$:
\begin{equation}
\tilde{Y}_{fg} = \mathcal{A}_{fg}(\text{Maxpool}(T,T_{ig},T_d))
\end{equation}

For coarse-grained calibration, predictions $\tilde{Y}_{cg}$ are derived by max-pooling only the outputs of the inter-grained encoder ($E_{ig}$) and coarse-grained decoder ($E_d$), considering the larger domain gap for coarse-grained VLPMs pretrained on natural scenes. Additionally, the coarse-grained adaptor employs a lightweight transformer decoder to better integrate coarse-grained features with the user query ($Q$):
\begin{equation}
\tilde{Y}_{cg} = \mathcal{A}_{cg}(\text{Maxpool}(E_{ig},E_d) \parallel Q)
\end{equation}

Through this two-stage adaptation—first with synthetic data, then with guided fine-tuning—the framework $\mathcal{F}$ becomes better aligned with domain-specific structure, semantics, and task requirements, achieving performance comparable to training on large-scale curated datasets.


\section{Environmental Setup}
\subsection{Datasets}
\paragraph{\textbf{Benchmark Datasets}} Two domain-specific and unstructured scanned document key information extraction datasets are utilized to evaluate the effectiveness of the SynJAC. 
\noindent\textbf{CORD} \citep{cord} provides multi-level annotations to support a range of task-specific or end-to-end printed/scanned (P) receipt understanding tasks. In line with previous document understanding frameworks \citep{layoutlmv2,layoutlmv3}, our focus lies on sequence tagging to identify the entity type of each textual token extracted from scanned receipts, including "\textit{store name}", "\textit{menu quantity}", and "\textit{void total}". 

\noindent\textbf{Form-NLU} \citep{formnlu}: delves into understanding layout structure (Task A) and extracting key information (Task B) from digital (D), printed (P), and handwritten (H) financial forms obtained from Australian Stock Exchange filings. This paper specifically focuses on Task B, which supplies ground truth bounding boxes of form semantic entities and query text (e.g., "\textit{Shareholder Name}", "\textit{Share Class}"), enabling the utilization of the proposed model to retrieve the target entity.

\paragraph{\textbf{Dataset Statistics}}
The detailed statistics of the datasets, including machine-generated synthetic data, are presented therein. For the FormNLU dataset, which consists of text-embedded forms parsable by PDF tools, entity counts correspond to text lines extracted using PDFMiner \footnote{\url{https://pypi.org/project/pdfminer/}}. In the case of the CORD dataset, which comprises scanned receipts, we employ PaddleOCR \footnote{\url{https://github.com/PaddlePaddle/PaddleOCR}} for text line extraction, yielding approximately 13,200 entities.

\begin{table*}[ht]
\centering
\renewcommand{\arraystretch}{1.2}
\setlength{\tabcolsep}{5pt}
\begin{adjustbox}{max width=\linewidth}
\begin{tabular}{l|ccc|c|c|c|c|c|cccc}
\hline
\multirow{2}{*}{\textbf{Dataset}} & \multicolumn{3}{c|}{\textbf{Split}} & \multirow{2}{*}{\textbf{Year}} & \multirow{2}{*}{\textbf{Domain}} & \multirow{2}{*}{\textbf{Task}} & \multirow{2}{*}{\textbf{Script}} & \multirow{2}{*}{\textbf{Lang.}} & \multicolumn{4}{c}{\textbf{Synthetic Dataset Size}} \\ 
\cline{2-4} \cline{10-13}
 & \textbf{Train} & \textbf{Val} & \textbf{Test} &  &  &  &  &  & \textbf{\# IMG} & \textbf{\# Entities} & \textbf{\# QA} & \textbf{\# Cat} \\ 
\hline
FormNLU & 535 & 76 & 50/50 & 2023 & Financial & Entity Retrieval & P/H & En. & 535 & 103{,}866 & 15{,}278 & N/A \\ 
CORD & 800 & 100 & 100 & 2019 & Receipt & Sequence Tagging & P & En. & 800 & 13{,}200 & N/A & 40 \\ 
\hline
\end{tabular}
\end{adjustbox}
\caption{
\textbf{Overview of adopted datasets and their synthetic variants.} Each dataset includes train/validation/test splits, domain and task types, and statistics on synthetic annotations such as image, entity, QA, and category counts.
}
\label{table:dataset_overview}
\end{table*}





\subsection{Baselines}
We employ a variety of pretrained backbones from both fine-grained and entity-level frameworks to encode multi-granularity features. \textbf{1) Fine-grained Baselines}
We utilize three recently proposed fine-grained document understanding models: LayoutLMv3 \citep{layoutlmv3}, LiLT \citep{lilt}, and UDop \citep{udop}, which leverage multimodal information pretrained on general document collections, like IIT-CDIP \citep{lewis2006building}, to perform key information extraction through sequence tagging tasks, achieving state-of-the-art performance when fully trained on benchmark datasets.
\textbf{2) Coarse-grained Baselines}
For entity-level document understanding, we include RoI-based VLPMs such as LXMERT \citep{lxmert} and VisualBERT \citep{visualbert} as baselines for entity retrieval. After properly fine-tuning those models on the well-annotated dataset, they can perform well on VRD QA or KIE tasks. We follow the configurations of baseline models for both token and entity levels as specified in \citep{layoutlmv3, lilt, udop, formnlu}. 
\textbf{3) Zero-shot LLMs and MLLMs}
LLMs and MLLMs have shown impressive zero-shot performance across diverse domains. To assess their capabilities on VRDU tasks, we evaluate GPT-3.5 and GPT-4, leading closed-source models for mono-modality and multimodal tasks. For open-source models, we select QWen-VL \citep{qwen} (pretraining-based), LLAVA-1.5 \citep{llava1.5} (instruct-tuned), and BLIP-3 \citep{blip3} (pretrained with instruct-tuning) based on their distinct training strategies. 

\subsection{Implementation Details}
\label{app:implementation_detail}

\begin{wraptable}{r}{0.5\linewidth}
\vspace{-10pt}
\centering
\renewcommand{\arraystretch}{1.15}
\setlength{\tabcolsep}{4pt}
\begin{adjustbox}{max width=\linewidth}
\begin{tabular}{p{3.4cm} p{6.2cm}}
\toprule
\rowcolor{gray!10}
\textbf{Component} & \textbf{Description} \\ 
\midrule
\textbf{Learning Rate} & $2\times10^{-5}$ \\
\textbf{Optimizer} & AdamW \\
\textbf{Batch Size} & 2 \\[2pt]
\textbf{Training Epochs} & 
\begin{tabular}[t]{@{}l@{}} 
CORD: 60 epochs \\ 
FormNLU: 15 epochs 
\end{tabular} \\[3pt]
\textbf{Hardware Setup} & Single NVIDIA V100 GPU (16GB) \\[2pt]
\begin{tabular}[t]{@{}l@{}} \textbf{Training Time} \\ (per epoch) \end{tabular} & 
\begin{tabular}[t]{@{}l@{}} 
Domain Adaptation: $\sim$10 min \\ 
Domain Calibration: $\sim$3 min 
\end{tabular} \\ 
\bottomrule
\end{tabular}
\end{adjustbox}
\caption{\revise{\textbf{Experimental Configuration.} Summary of the key hyperparameter and training setup used in the proposed SynJAC framework.}}
\label{tab:exp_config}
\vspace{-5pt}
\end{wraptable}
We follow the baseline configurations for both token-level and entity-level models as specified in \citep{layoutlmv3, lilt, udop, formnlu}. LayoutLMv3 and LXMERT serve as the fine-grained ($\mathcal{E}_{fg}$) and coarse-grained ($\mathcal{E}_{cg}$) general-domain pretrained encoders, respectively, chosen for their strong performance demonstrated in prior work \cite{m3vrd}. Our architecture employs a six-layer transformer encoder with a hidden size of 768 as the joint-grained encoder ($\mathcal{E}_{jg}$), along with two additional six-layer transformer decoders—also with a hidden size of 768—which function as the fine-grained ($\mathcal{D}_{fg}$) and coarse-grained ($\mathcal{D}_{cg}$) decoders.
\revise{Table~\ref{tab:exp_config} summarizes the training-related settings, including hyperparameter, optimizer choices, the number of training epochs for domain calibration, and the average runtime required to obtain results. Regarding the training epochs of various domain adaptation strategies, we conduct the ablation testing in Table~\ref{tab:ablation_results} to show the effects of training epochs and combination strategies}. The full model architecture and the number of trainable parameters are presented in Table~\ref{tab:num_parameters}, and the efficiency of the proposed SynJAC framework is reported in Table~\ref{tab:time_effciency}.

\subsection{\revise{Synthetic Data Generation Setup}}
\revise{In Table~\ref{tab:synthetic_tasks_templates}, we present prompt examples for \textit{Synthetic Tag Generation} and \textit{Synthetic Inquiry Generation}. Each task includes a descriptive instruction and an in-context example to ensure the LLM correctly understands the objective. We use GPT-4o-turbo for generation, with a temperature of 1 and a maximum output length of 2048 tokens. Post-processing is task-specific: all outputs are first parsed and validated against the expected format. For synthetic tag generation, we further verify that the number of tags matches the token list and that all tags belong to predefined entity types; mismatches are added to a hold set. For inquiry generation, the LLM evaluates if the questions are meaningful and if answers are relevant—only examples with affirmative checks are retained. Documents with fewer than ten valid inquiries are also moved to the hold set for regeneration.}
 
\begin{table}[h!]
\centering
\footnotesize
\begin{adjustbox}{max width=\linewidth}

\begin{tabular}{M{0.14\linewidth} M{0.3\linewidth} M{0.25\linewidth} M{0.25\linewidth}}
\toprule
\textbf{Type} & \textbf{Instruction} & \multicolumn{2}{c}{\textbf{In-Context Example}} \\
\midrule

\textbf{Synthetic Tag Generation} &
Given a list of tokens extracted from a domain document  
and a list of entity types, generate BIO-tagged labels  
that align exactly with the token sequence. &
\textbf{Entity Types:} Date, Name, Location 
\textbf{Example Tokens:}  
["The", "Previous", "notice", "was", "dated", "26", "/", "04", "/", "2017"]  
&
\textbf{Expected Outputs:}  
["O", "O", "O", "O", "O",  
 "B-Date", "I-Date", "I-Date", "I-Date", "I-Date"]  
\\
\midrule

\textbf{Synthetic Inquiry Generation} &
Given document context, generate <N> questions with  
answers that must be *directly extractable* from  
the provided context. &
\textbf{Context:}  
“Form 604 Corporations Act 2001…  
Issued securities: Ordinary Shares …”  
 &
\textbf{Expected Outputs:} 
\textbf{Q1:} What is the type of the share class?  
\textbf{A1:} Ordinary Shares  

\textbf{Q2:} XXX?  
\textbf{A2:} XXX
\\
\bottomrule

\end{tabular}
\end{adjustbox}

\caption{Prompt Formats for Synthetic Data Generation.}
\label{tab:synthetic_tasks_templates}
\end{table}

\section{Results and Discussion}

\subsection{Synthetic Data Analysis}
\label{sec:syn_data}
\begin{figure*}[thb]
    \hspace*{0cm}
     \centering
     \hspace*{-0.5em}
     \begin{subfigure}[b]{0.24\textwidth}
         \centering
         \includegraphics[height=2.6cm]{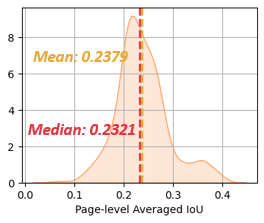}
         \caption{FormNLU \textit{Syn-Struct}}

         \label{fig:bbox_noise_formnlu}
     \end{subfigure}
     \hspace*{-0.3em}
     \begin{subfigure}[b]{0.24\textwidth}
         \centering
         \includegraphics[height=2.6cm]{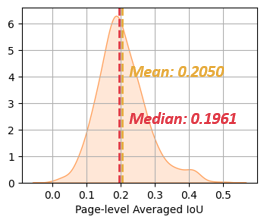}
                   \caption{CORD \textit{Syn-Struct}}
        \label{fig:bbox_noise}

     \end{subfigure}
     \hspace*{-0.3cm}
     \centering
     \hspace*{-0.3em}
     \begin{subfigure}[b]{0.24\textwidth}
         \centering
         \includegraphics[height=2.6cm]{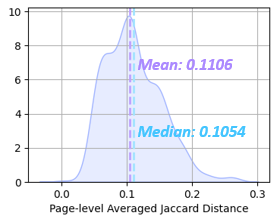}
          \caption{FromNLU \textit{Syn-Text}}
         \label{fig:text_noise_formnlu}
     \end{subfigure}
     \hspace*{-0.5em}
     \begin{subfigure}[b]{0.24\textwidth}
         \centering
         \includegraphics[height=2.6cm]{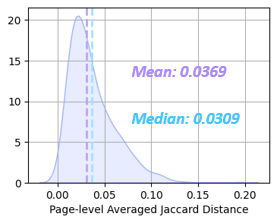}
           \caption{CORD \textit{Syn-Text}}

         \label{fig:text_noise}
     \end{subfigure}

        \caption{Off-the-shelf-tool analysis. Synthetic-Structure (\textit{Syn-Struct}) and Synthetic-Text (\textit{Syn-Text}).}
        \label{fig:tool_noise_distribution}
         \vspace*{-0.5em}
\end{figure*}


We firstly analyse the distribution characteristics of synthetic annotations generated by off-the-shelf tools, focusing on two primary types: \textbf{1) Layout structure variations} arise from inaccuracies in the regions of document semantic entities extracted by document parsing tools. However, text content variations result from improperly grouped words and mis-recognised text during the parsing process.
From Figures~\ref{fig:bbox_noise} and \ref{fig:bbox_noise_formnlu}, most documents exhibit mismatches in layout structures, with the average Intersection over Union (IoU) between detected entities and ground truth entities falling below 0.3 in both datasets. \textbf{2) Text content variations} exhibit even lower Jaccard similarities, dropping below 0.2 for Form-NLU and 0.1 for CORD. Errors in entity detection can propagate during text recognition, resulting in a larger distribution gap between extracted text sequences and the ground truth. Compared to text-embedded source files that can be processed by PDF parsing tools like PDFMiner, scanned documents processed by OCR tools tend to introduce even more variations, further complicating the adaptation of models to these documents. However, as demonstrated by the experimental results, despite the presence of noise in the synthetically generated data, it effectively enables the general-domain pretrained framework to better comprehend the target domain when integrated with the joint-grained model and domain adaptation strategies. 

\subsection{\revise{SynJAC Overall Performance Analysis}} \label{sec:overall_performance}

\begin{wraptable}{r}{0.5\linewidth}
\centering
\small
\begin{adjustbox}{max width=\linewidth}
\begin{tabular}{c|cc|c|c}
\toprule
\multirow{2}{*}{\textbf{Entity Level}} & \multicolumn{2}{c|}{\textbf{FormNLU}} & \multirow{2}{*}{\textbf{Token Level}} & \multirow{2}{*}{\textbf{CORD}} \\
\cmidrule{2-3}
 & \textbf{P} & \textbf{H} & & \\
\midrule
\multicolumn{5}{c}{\textbf{Full Training Set (Manually Annotated Training Set)}} \\
\midrule
Transformer & 88.62 & 74.06 & LayoutLMv3 & 96.56 \\
VisualBERT  & 85.90 & 70.14 & LiLT       & 96.07 \\
LXMERT      & \textbf{94.15} & \textbf{82.80} & UDOP      & \textbf{97.58} \\
\midrule

\multicolumn{5}{c}{\textbf{Tuning in Guidance Set ($\mathbb{D}_g$)}} \\
\midrule
Transformer & 72.82 & 60.30 & LayoutLMv3 & 87.08 \\
VisualBERT  & 46.48 & 48.41 & LiLT       & 86.74 \\
LXMERT      & 81.21 & 64.66 & UDOP       & 80.88 \\
\midrule
Vanilla         & 89.60 & 85.76 & Vanilla        & 87.48 \\
+ L2V           & 90.60 & 87.60 & + L2V          & 88.11 \\
+ SDS           & 91.11 & 88.78 & + SDS          & 89.08 \\
+ SIT           & 90.77 & 87.94 & + SST          & 88.83 \\
+ SIT + SDS     & \textbf{92.62} & \textbf{89.11} & + SST + SDS    & \textbf{90.25} \\
\bottomrule
\end{tabular}
\end{adjustbox}
\caption{Performance of various frameworks using full manually annotated training set and guidance set. Bold values indicate the best in each group.}
\vspace{-5pt}
\label{tab:overall}
\end{wraptable}
\paragraph{\textbf{Overall Performance Analysis}}
Table~\ref{tab:overall} presents the performance of various model configurations, demonstrating the effectiveness of the proposed domain adaptation methods in capturing domain knowledge. Firstly, the results show that integrating fine and coarse-grained information outperforms mono-grained baselines, boosting downstream task performance.
We note that incorporating fine-grained features significantly enhanced coarse-grained representation in FormNLU, with a performance gain of approximately 8\% for the printed and 21\% for the handwritten sets. All domain adaptation methods, including the novel L2V positional features, improved performance. Detailed analyses are in subsequent sections.

\begin{table*}[h]
    \centering
    \begin{adjustbox}{max width =\linewidth}
    \begin{tabular}{l|c|c|c|c|c|c|c|c|l|c|c|c|c}
        \hline
        \multirow{3}{*}{\textbf{Entity Level}} & \multicolumn{8}{c|}{\textbf{FormNLU}} &\multirow{3}{*}{\textbf{Token Level}} & \multicolumn{4}{c}{\textbf{CORD}} \\
        
        \cline{2-9} 
        \cline{11-14}
        
         &\multicolumn{2}{c|}{\textbf{\textit{com\_id}}}&  \multicolumn{2}{c|}{\textbf{\textit{ntc\_dt}}} &  \multicolumn{2}{c|}{\textbf{\textit{gvn\_dt}}} &  \multicolumn{2}{c|}{\textbf{\textit{prv\_pct}}} & &\multirow{2}{*}{\textbf{\textit{sc}}} & \multirow{2}{*}{\textbf{\textit{up}}} & \multirow{2}{*}{\textbf{\textit{ccp}}} & \multirow{2}{*}{\textbf{\textit{setc}}}   \\
         
         \cline{2-9} 
         & P & H & P & H & P & H & P & H & & & & & \\
        \hline
       LXMERT & 45.83 & 30.00 & 72.00 & 69.39 & 78.00 & 83.67 & 98.00 & 67.35 & LayoutLMv3 & 55.17 & 93.53 & 85.71 & 82.54 \\
        Joint-grained & 50.00 & \textbf{88.00 }& 66.00 & 18.37 & 92.00 & 79.80 & \textbf{100.00} & 89.80 & Joint-grained & 73.33 & 85.51 & 91.67 & 76.92 \\
        + L2V & 66.67 & 72.00 & 72.00 & 61.22 & 88.00 & \textbf{95.92} & \textbf{100.00} & 95.92 & + L2V & 64.29 & 94.12 & 84.62 & 82.54  \\
        + SDS & \textbf{79.17} & \textbf{88.00} & 66.00 & 61.22 & 88.00 & 89.80 & \textbf{100.00} & 95.92 &+ SDS & 80.00 & 94.89 & \textbf{100.00} & 89.23 \\
        + FST & 62.50 & 78.00 & 72.00 & 67.35 & 90.00 & 85.71 & \textbf{100.00} & \textbf{100.00} & + SST & \textbf{84.85} & 91.43 & 80.00 & 80.65  \\
         + FST + SDS & \textbf{79.17} & 78.00 & \textbf{80.00} & \textbf{81.63} & \textbf{92.00} & 85.71 & 96.00 & 95.92  & + SST + SDS & 64.29 & \textbf{97.06} & 88.89 & \textbf{90.32}\\
        \hline
        \multicolumn{14}{c}{\small Note: `\textit{com\_id}` = company identifier (ACN/ARSN), `\textit{ntc\_dt}` = notice date `\textit{gvn\_dt}` = notice given to company date, `\textit{prv\_pct}` = previous voting power}\\
        \multicolumn{14}{c}{\small `\textit{sc}` = subtotal count, `\textit{up}` = unit price, `\textit{ccp}` = credit card price, `\textit{setc}` = subtotal others}\\
        \hline
    \end{tabular}
    \end{adjustbox}
    \caption{Selective breakdown results of performance across representative categories.}
            \label{tab:breakdown}
\end{table*}

\paragraph{\textbf{\revise{Representative Category Breakdown Analysis}}}
\label{sec:selected_breakdown}
Table~\ref{tab:breakdown} compares performance across various information categories, highlighting the benefits of the the proposed joint-grained framework and domain adaptation approaches in generating comprehensive representations. This framework enriches entity semantics and token structures, leading to notable improvements—such as a 58\% increase in ``\textit{com\_id}" in FormNLU-H and an 18\% increase in "\textit{sc}" in CORD. While L2V enhances feature representation overall, it may introduce inconsistencies in flexible layout categories, like handwritten `\textit{com\_id}" in FormNLU. The proposed methods, especially SDS, consistently show robust improvements across most categories, demonstrating their effectiveness in capturing domain-aware knowledge. Although leveraging LLM-generated tags (SST) or QA pairs (SIT) boosts performance, it may lead to occasional instability. For example, combining SDS with SST or SIT improve specific categories but may yield lower results in others—such as a 20\% decrease in CORD's "\textit{sc}" when using SDS+SST compared to SST. \revise{These fluctuations are likely caused by noise introduced during the synthetic data generation stage. As document parsing techniques and MLLM-based synthetic data generation frameworks continue to advance, the quality and consistency of synthetic supervision are expected to improve, thereby reducing such errors. Nevertheless, despite occasional instability, the majority of results in Table~\ref{tab:overall} and Table~\ref{tab:breakdown} show that our training strategies effectively narrow the domain gap and deliver substantial performance gains with limited computational cost.}

\subsection{Results with Stepped Training Ratios}
\label{sec:few_shot}
\begin{figure}[thb]
    \hspace*{0cm}
     \centering
     \hspace*{-0.8em}
     \begin{subfigure}[b]{0.34\textwidth}
         \centering
         \includegraphics[height=4cm]{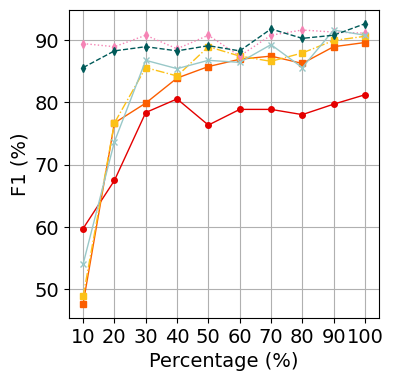}
          \caption{P}
         \label{fig:few_shot_cord}
     \end{subfigure}
     \begin{subfigure}[b]{0.3\textwidth}
         \centering
         \includegraphics[height=4cm]{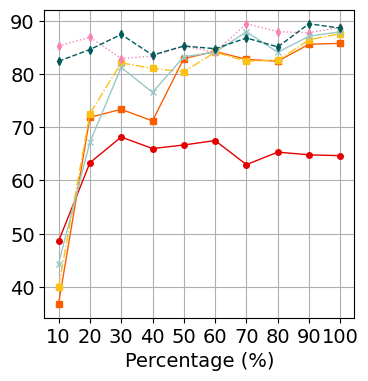}
          \label{fig:few_shot_handwritten}
\caption{H}
         \label{fig:few_shot_printed}
     \end{subfigure}
     \begin{subfigure}[b]{0.3\textwidth}
         \centering
         \includegraphics[height=4cm]{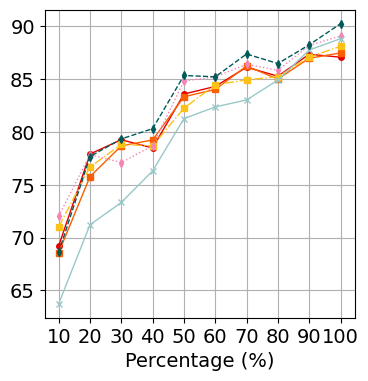}
         \caption{CORD}
     \end{subfigure}
     \begin{subfigure}[b]{\textwidth}
         \centering
         \includegraphics[height=0.35cm]{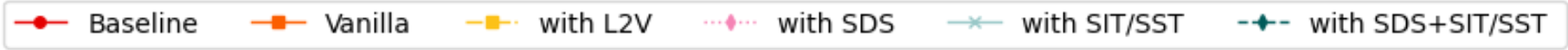}
     \end{subfigure}
        \caption{Performance of SynJAC with stepped training set ratios on three test sets.}
        \label{fig:few_shot}
        \vspace{-1em}

\end{figure}

\paragraph{\textbf{Few-shot Testing}} We evaluated the robustness of our methods with varying amounts of annotated data of the guidances set $\mathbb{D}_g$, using training sizes from 10\% to 100\%. As shown in Table~\ref{tab:overall}, applying domain adaptation consistently outperformed non-adapted baselines by leveraging domain-specific information from the synthetic annotated $\mathbb{D}_s$, although performance sensitivity varied across different tasks and training sizes.
For the coarse-grained FormNLU, both printed (P) and handwritten (H) test sets improved as training sizes increased. Without domain adaptation, performance was poor in few-shot scenarios. With just 10\% of $\mathbb{D}_g$, applying SDS achieved over 80\% accuracy on both P and H sets, demonstrating its ability to capture domain-specific structural information.
For fine-grained KIE in CORD, incorporating coarse-grained information improved performance across training sizes. SDS consistently outperformed other configurations, effectively utilizing synthetic structural information from $\mathbb{D}_s$. However, SIT and SST underperformed in few-shot settings, likely due to reliance on synthetic LLM-generated samples that need more data to bridge distribution gaps.

\begin{wraptable}{r}{0.5\linewidth}
\vspace{-1em}
\centering
\small 
\renewcommand{\arraystretch}{0.85} 
\begin{adjustbox}{max width=\linewidth}
\begin{tabular}{lll|ll}
\toprule
\multicolumn{3}{c|}{\textbf{FormNLU}} & \multicolumn{2}{c}{\textbf{CORD}} \\
\midrule
\textbf{Config} & \textbf{P} & \textbf{H} & \textbf{Config} & \textbf{Test} \\
\midrule
Baseline & 1.67 & 0.5 & Baseline & 0 \\
Joint-grained & 0 & 0 & Joint-grained & 0 \\
+ L2V & 0 & 0 & + L2V & 0 \\
+ SDS & \textbf{87.42} & \textbf{81.74} & + SDS & 0.05 \\
+ SIT & 5.7 & 0.17 & + SST & 0.25 \\
+ SIT + SDS & 47.65 & 44.22 & + SST + SDS & \textbf{4.21} \\
\bottomrule
\end{tabular}
\end{adjustbox}
\caption{Zero-shot performances on various configurations.}
\label{tab:zero_shot}
\end{wraptable}

\paragraph{\textbf{Zero-shot Testing}} We evaluated zero-shot performance (Table~\ref{tab:zero_shot}) to assess domain knowledge infusion of diverse domain adaptation strategies. SDS effectively distilled structural knowledge from $\mathbb{D}_s$, achieving 87.42\% on FormNLU (printed) and 81.74\% (handwritten). In contrast, SIT showed minor improvements on the printed set but decreased on the handwritten set, possibly due to the distribution gap between digital-born QA pairs from $\mathbb{D}_g$ and handwritten tests.
For CORD, domain adaptation had less impact than entity-level tasks, as the joint-grained framework benefits entity representations more than token representations. 

\begin{wraptable}{r}{0.55\linewidth}
\centering
\vspace{-1em}
\small
\renewcommand{\arraystretch}{0.9} 
\begin{adjustbox}{max width=\linewidth}
\begin{tabular}{@{}lcc|lc@{}}
\toprule
\multicolumn{3}{c|}{\textbf{FormNLU}} & \multicolumn{2}{c}{\textbf{CORD}} \\
\midrule
\textbf{Config} & \textbf{P} & \textbf{H} & \textbf{Config} & \textbf{Test} \\
\midrule
\multicolumn{5}{>{\columncolor{gray!20}}c}{Domain Adaptation Ablation Testing} \\
\midrule
\revise{Vanilla} & \revise{89.60} & \revise{85.76} & \revise{Vanilla} & \revise{88.45} \\

\midrule
SDS (ep. 1) & \underline{91.11} & \underline{88.78} & SDS (ep. 1) & 88.45 \\
SDS (ep. 2) & 89.93 & 86.60 & SDS (ep. 2) & \underline{89.08} \\
SDS (ep. 3) & \underline{91.11} & 84.42 & SDS (ep. 3) & 87.35  \\
\midrule
SIT (ep. 1) & \underline{90.94} & \underline{87.77} & SST (ep. 1) & \underline{88.83}  \\
SIT (ep. 2) & 86.91 & 83.75 & SST (ep. 2) & 87.54 \\
SIT (ep. 3) & 86.07 & 81.41 & SST (ep. 3) & 85.71 \\

\midrule
SDS+SIT (ep. 1) & 91.11 & \underline{89.11} & SDS+SST (ep. 1) & 86.95  \\
SDS+SIT (ep. 2) & \underline{92.62} & 88.61 & SDS+SST (ep. 2) & \underline{90.25} \\
SDS+SIT (ep. 3) & 87.58 & 83.92 & SDS+SST (ep. 3) & 87.49  \\

\midrule
SDS with L2V & \underline{91.11} & \underline{89.11} & SDS with L2V & \underline{89.08} \\
SDS without L2V & 89.26 & 84.25 & SDS without L2V & 87.57 \\
SIT with L2V & 90.94 & 87.77 & SST with L2V & 88.83  \\
SIT without L2V & 85.91 & 87.94 & SST without L2V & 87.19  \\
\midrule
\multicolumn{5}{>{\columncolor{gray!20}}c}{Domain Calibration Ablation Testing} \\
\midrule
SDS Frozen & 91.11 &  \underline{88.78} & SDS Frozen & \underline{89.08}  \\
SDS Unfrozen & 91.61 & 85.59 & SDS Unfrozen & 86.91  \\
SDS+SIT Frozen & \underline{92.62} & 85.59 &  SDS+SST Frozen & \underline{90.25}  \\
SDS+SIT Unfrozen & 88.59 & 85.93 & SDS+SST Unfrozen & 86.64 \\

\midrule
$E_{d}$ & \underline{92.64} & 86.60  & $T_{d}$  & 88.67\\
$E_{d} + E_{ig}$ & 92.62 & \underline{89.11} & $T_{d}+T_{ig}$ & 88.98\\
$E_{d} + E_{ig} + E_{cg}$ & 91.78 & 88.21 & $T_{d}+T_{ig}+T_{fg}$ & \underline{90.25}  \\
\midrule

Max Pooling & \underline{92.62} & \underline{89.11} & Max Pooling & \underline{90.25} \\
Mean Pooling & 91.23 & 86.44 & Mean Pooling & 87.57 \\
Min Pooling & 89.78 & 85.59 & Min Pooling & 87.63  \\
\revise{Concatenation} & \revise{92.02} & \revise{87.87} & \revise{Concatenation} & \revise{87.61} \\

\bottomrule
\end{tabular}
\end{adjustbox}
\caption{Ablation studies. The best performed configuration of each group is underlined.}
\label{tab:ablation_results}
\vspace{-1em}

\end{wraptable}
\subsection{\revise{Ablation Studies: Domain Adaptation Stage}}
\label{sec:ablation_adaptation}
\revise{In the upper part of Table~\ref{tab:ablation_results}, we compare various domain adaptation strategies and demonstrate the effectiveness of integrating L2V during domain adaptation.}

\label{sec:ablation_study_da}
\paragraph{\textbf{Effects of Domain Adaptation Training Epochs}}
We observed that varying the number of training epochs (ep.) for different domain adaptation strategies impacts final results in Table~\ref{tab:ablation_results}. Insufficient training may limit the model’s ability to acquire domain-specific information. For instance, training the SDS+SST method for just one epoch on the CORD dataset yields about 2.5\% lower performance than two epochs. Conversely, increasing training epochs can cause the model distribution to shift closer to $\mathbb{D}_s$, but further away from $\mathbb{D}_g$, as seen with SDS+SIT on FUNSD, where three epochs caused ~2.5\% and 5\% drops on sets P and R, respectively. \revise{Although the optimal number of epochs depends on the dataset and task—requiring careful tuning and further investigation, we observe that even a small number of training epochs can yield substantial performance improvements. Additionally, developing a principled mechanism to automatically determine the appropriate number of domain-adaptation epochs, potentially informed by synthetic-label noise estimation or validation-based stability criteria, is an important direction for future work.}

\paragraph{\textbf{Effects of L2V Positional Embedding}}
We evaluated the impact of the L2V positional embedding on domain adaptation methods. As shown in Table~\ref{tab:ablation_results}, removing L2V led to an approximate 2\% performance drop. This suggests that L2V enhances positional awareness in fine and coarse-grained representations, consistently contributing to better document understanding ability. Notably, it could lead to a promising increase on FormNLU handwritten (FormNLU-H) set (around 5\%) which demonstrates the effectiveness the L2V embedding could enhance the layout representation to improve the framework generalisability. 

\subsection{\revise{Ablation Studies: Domain Calibration Stage}}
\label{sec:ablation_calibration}


\paragraph{\textbf{Effects of Freezing Inter-grained Encoder}} 
Freezing the parameters of the inter-grained encoder $\mathcal{E}_{ig}$ after applying SDS was effective in preserving domain knowledge acquired from $\mathbb{D}_s$. This strategy maintained structural and semantic representations while mitigating overfitting to the guidance set, thereby enhancing guidance-based domain calibration. As evidenced in Table~\ref{tab:ablation_results}, unfreezing led to performance degradation; for instance, SDS+SIT on FormNLU-P decreased to 88.58\% without freezing. 

\paragraph{\textbf{Effects of Latent Feature Combination}} Features extracted from different stages capture varied knowledge, shaped by the backbone architecture, input representations, and training objectives. For example, features from a general-domain pretrained backbone—used as coarse-grained ($T_{fg}$) and fine-grained ($E_{cg}$) representations—encode broad world knowledge. After structural domain shifting (SDS), inter-grained features ($T_{ig}$ and $E_{ig}$) are further enriched by inter- and intra-grained signals from the synthetic dataset $\mathbb{D}_s$. Subsequently, task-specific features ($T_{d}$ and $E_{d}$) are refined with the manually annotated guidance set $\mathbb{D}_g$ to support domain adaptation.
To evaluate the contribution of each feature stage, we present ablation results in Table~\ref{tab:ablation_results}. Relying solely on task-specific features ($T_{d}$ and $E_{d}$) often leads to overfitting on the guidance set, particularly in the FormNLU handwritten subset (H). In contrast, integrating features across stages improves robustness. Notably, coarse-grained features from general-domain pretrained encoders \cite{visualbert,lxmert}, such as those trained on natural scenes, underperform, likely due to a significant domain gap with document-centric tasks.

\paragraph{\textbf{Effects of Various Pooling Methods}} To effectively leverage representations from diverse stages, we investigate various pooling strategies, including max pooling, mean pooling, and min pooling,  and concatenation following a linear project layer for aggregating multi-stage features. As shown in Table~\ref{tab:ablation_results}, max pooling consistently outperforms the other methods across both datasets. This indicates that max pooling is more effective in capturing the most informative signals from multi-stage representations. Its superior performance suggests an enhanced ability to balance knowledge learned from synthetic and curated datasets, thereby mitigating noise in domain adaptation and reducing overfitting during domain calibration.


\subsection{Robustness Analysis}
\label{app:robustness_analysis}
\begin{wraptable}{r}{0.57\linewidth}
\vspace{-1em}

\centering
\renewcommand{\arraystretch}{0.9} 
\begin{adjustbox}{max width=\linewidth}
\begin{tabular}{l|ccc|ccc}
\toprule
\multirow{2}{*}{Model} & \multicolumn{3}{c|}{$X \sim N(0,1)$, $y\neq \hat{y}$} & \multicolumn{3}{c}{$X \sim N(0,1)$, $\hat{y} = \emptyset $} \\ \cmidrule{2-7} 
                       & $P_2$ & $P_{1.5}$ & $P_1$  & $P_2$ & $P_{1.5}$ & $P_1$ \\ \midrule
Baseline & \underline{86.08} & \underline{82.65} & 74.83 & 85.58 & 82.09 & 75.20 \\ 
Joint-grained  & 85.47 & \textbf{82.81} & 74.45 & \underline{86.21} & \underline{82.79} & \underline{76.40} \\ 
+SDS & 84.28 & 81.79 & 74.62 & 85.78 & 80.19 & \textbf{76.82} \\ 
+SST & 85.70 & 81.96 & \underline{75.73} & 84.36 & 81.99 & 75.80 \\ 
+SDS+SST & \textbf{87.20} & 82.26 & \textbf{76.23} & \textbf{86.32} & \textbf{82.89} & 75.52 \\ 
\bottomrule
\end{tabular}
\end{adjustbox}
\caption{Performance comparison of models under different types of synthetic annotation label (incorrect and incomplete) across varying synthesis ratios.}
\label{tab:robust}
\vspace{-0.5em}

\end{wraptable}
\paragraph{\textbf{Synthetic Data Quality Analysis}} To evaluate the robustness of the proposed framework and domain adaptation strategies, synthetic label noise was introduced into the guidance set \(\mathbb{D}_g\) of the CORD dataset. Instances were randomly selected using a normal distribution, \(X \sim \mathcal{N}(0,1)\), and their ground truth labels \(y\) were replaced with randomly chosen labels \(\hat{y}\) from the label space \(Y\) or assigned "Unknown" (\(\emptyset\)). By controlling the parameter \(\lambda\), the proportion of noisy instances was adjusted to \(P(|X| > \lambda) = P_{\lambda}\), enabling an in-depth analysis of the framework's ability to handle varying levels of label corruption.
As shown in Table~\ref{tab:robust}, the joint-grained framework consistently demonstrates superior robustness compared to the baseline in both incorrect and incomplete label scenarios. Its integration of coarse-grained information significantly mitigates the negative impact of noisy or missing labels. Domain adaptation strategies further enhance performance, illustrating the framework's capability to adapt to challenging, label-deficient conditions in real-world applications.

\paragraph{\textbf{Synthetic Data Quantity Analysis}}

\begin{wraptable}{r}{0.5\linewidth}

\centering
\begin{adjustbox}{max width =\linewidth}
\begin{tabular}{l|c|c|l|c}
\toprule
\multirow{2}{*}{\textbf{Config.}} & \multicolumn{2}{c|}{\textbf{Form NLU}} & \multirow{2}{*}{\textbf{Config.}} & \multirow{2}{*}{\textbf{CORD}} \\ \cline{2-3}
& \textbf{P} & \textbf{H}  &  & \\

\midrule
No DW & 89.60 & 85.76 & No DW & 88.11 \\ 
\midrule
½ SDS & 90.60 & \underline{86.93} & ½ SDS & \underline{89.27} \\ 
½ SIT & \underline{91.28} & 85.76 & ½ SST & 87.93 \\ 
½ SDS+SIT & 90.60& 85.59 & ½ SDS+SST & 88.25 \\ 
\midrule
SDS & 91.11 & \textbf{88.78} & SDS & 89.08 \\ 
SIT & 90.77 & 87.94 & SST & 88.83 \\ 
SDS+SIT & \textbf{92.62} & 88.61 & SDS+SST & \textbf{90.25} \\ 
\bottomrule
\end{tabular}

\end{adjustbox}

\caption{Effects of changing the size of synthetic annotated set $\mathbb{D}_s$}
\label{tab:noise_set_quantity}
\vspace{-0.5em}

\end{wraptable}

In practical settings, the number of synthetic document collections is often limited due to domain-specific constraints.
To evaluate the impact of varying $\mathbb{D}_s$ sizes, we analysed how performance changes with different synthetic set sizes, as shown in Table~\ref{tab:noise_set_quantity}  to demonstrate the effectiveness of the proposed framework. Generally, increasing $\mathbb{D}_s$ improves model performance during fine-tuning on $\mathbb{D}_g$. Domain adaptation methods that address structural domain shifts are less sensitive to $\mathbb{D}_s$ size, while methods like synthetic inquiry tuning and sequence tagging are more affected. This indicates that even a limited amount of synthetic structural information can effectively bridge domain gaps, though a larger $\mathbb{D}_s$ size further strengthens model robustness and overall performance.

\subsection{Comparison with LLMs/MLLMs}


\begin{wraptable}{r}{0.57\linewidth}
\centering
\small
\begin{adjustbox}{max width =\linewidth}
\begin{tabular}{l|c|c|c|c|c|c}
\midrule
\multirow{2}{*}{ \textbf{Model} }& \multicolumn{2}{c|}{\textbf{FormNLU P}} & \multicolumn{2}{c|}{\textbf{FormNLU H}} & \multicolumn{2}{c}{\textbf{CORD*}} \\ 
\cmidrule{2-7}
 & \textbf{Time} & \textbf{F1} & \textbf{Time} & \textbf{F1} & \textbf{Time} & \textbf{ANLS} \\
\midrule
GPT-3.5 & 03:49 & 34.37 & 04:38 & 30.94 & 01:16 & 28.15* \\
GPT-4o ($P_t$) & 04:46 & 42.09 & 04:19 & 36.00 & 01:48 & 29.55* \\
\midrule
LLava ($P_{tv}$) & 52:54 & 9.79 & 60:58 & 7.82 & 10:23 & 37.98 \\
QWen ($P_{tv}$) & 1:36:00 & 9.84 & 1:58:00 & 8.43 & 18:13 & 37.58 \\
Blip3 ($P_{tv}$) & 36:06 & 12.62 & 35:24 & 11.67 & 10:12 & 43.73 \\
GPT-4o ($P_{tv}$) & 20:02 & 59.88 & 20:49 & 49.15 & 07:55 & 79.46* \\
\midrule
SynJAC-ZS & 03:37 & 87.42 & 03:31 & 81.74 & - & - \\

SynJAC-$\mathbb{D}_g$& 03:37 & \textbf{92.62} & 03:31 & \textbf{88.78} & 00:31 & \textbf{90.25} \\
\bottomrule
\end{tabular}
\end{adjustbox}
\caption{Performance between LLM/MLLMs and SynJAC. CORD* is adopted QA-style subset introduced by LayoutLLM.}

\label{tab:llm_testing}
\end{wraptable}

We evaluated state-of-the-art LLMs and MLLMs on VRDU tasks using diverse mono- and multi-modal prompts across various training checkpoints, representing their performance and efficiency with the SynJAC framework (Table~\ref{tab:llm_testing}). For the proprietary GPT-4o, two prompts were tested: a text-only prompt $P_t: \{K, C\}$, where $K$ denotes key text and $C$ the provided content, and a multimodal prompt $P_{tv}: \{K, C, I\}$, with $I$ representing the target form image. GPT-3.5 utilized only $P_t$, while open-source MLLMs employed $P_{tv}$ to leverage both textual and visual modalities. GPT-4o with $P_t$ surpassed GPT-3.5, and achieved an F1 score improvement of approximately 13\% with $P_{tv}$. Open-source MLLMs showed a clear performance gap compared to the GPT series. Although LLMs/MLLMs operate in a zero-shot setting and are thus not directly comparable to few-shot frameworks, the substantial performance differences suggest the potential for integrating our plug-and-play agent to achieve efficient, high-performing domain-specific KIE.

\begin{figure*}[h]
  \centering

  \begin{subfigure}[b]{\linewidth}
    \centering
    \includegraphics[width=\linewidth]{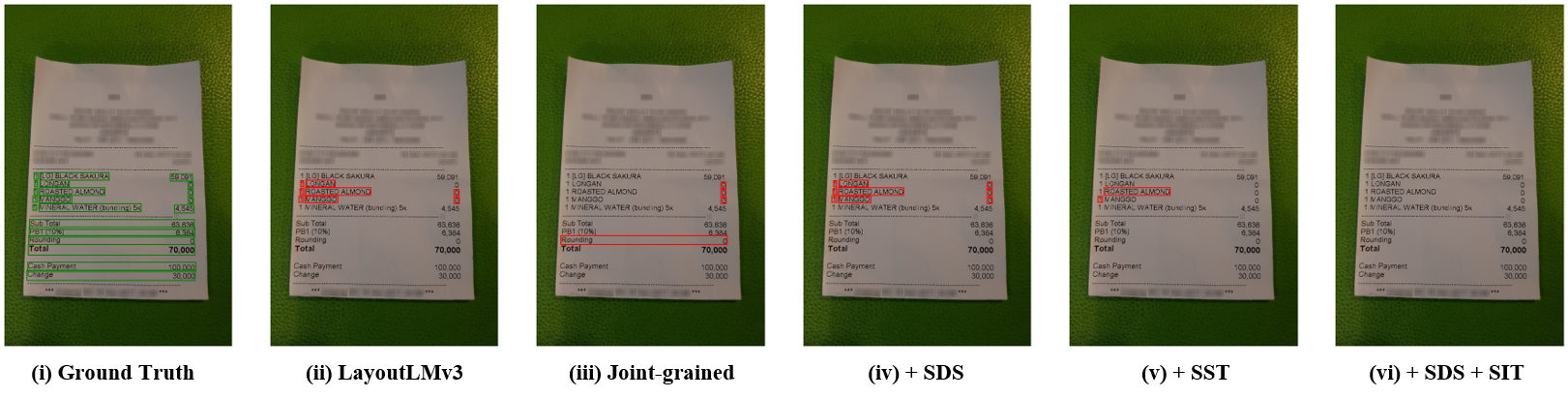}
    \caption{CORD dataset quantitively example.}
    \label{fig:case_study_main}
  \end{subfigure}

  \begin{subfigure}[b]{\linewidth}
    \centering
    \includegraphics[width=\linewidth]{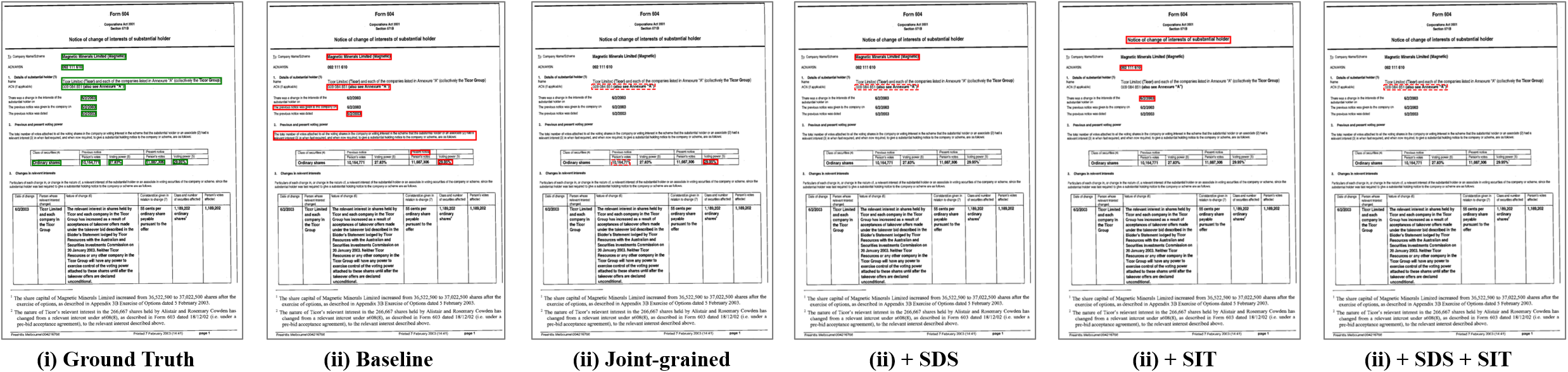}
    \caption{FormNLU printed dataset quantitively example.}
    \label{fig:case_study_formnlu_p1}
  \end{subfigure}
  \begin{subfigure}[b]{\linewidth}
    \centering
    \includegraphics[width=\linewidth]{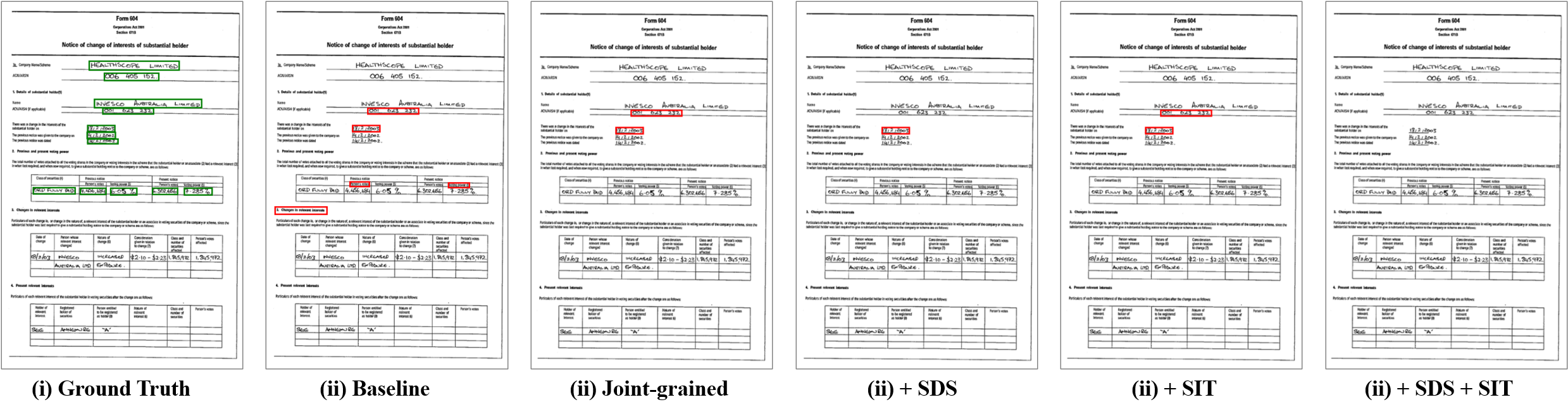}
    \caption{FormNLU handwritten dataset quantitively example.}
    \label{fig:case_study_formnlu_h4}
  \end{subfigure}
  \caption{(i) Ground truth key information highlighted in green. (ii)–(vi) Incorrect predictions marked with red rectangles under various configurations.}
  \label{fig:case_study_combined}
\end{figure*}
\section{Qualitative Analysis: Case Studies}

To qualitatively assess the effectiveness of \textit{SynJAC}, real-world examples of three subsets are shown in Figure~\ref{fig:case_study_combined}. Across all scenarios, the joint-grained framework consistently yields fewer incorrect predictions compared to baseline models, likely due to its ability to integrate inter-grained information and enhance document representations. Regarding domain adaptation, both structural (SDS) and task-oriented (SIT/SST) approaches contribute to reducing error rates. However, when applied individually, SDS or SIT/SST may underperform in certain cases, potentially due to noise in the synthetic data. Notably, combining both adaptation strategies consistently produces more accurate predictions. This demonstrates the proposed methods’ capacity to effectively leverage noisy, domain-specific annotations, ultimately improving performance in downstream tasks.

\section{Conclusion}

\revise{This paper introduced SynJAC, a synthetic-data-driven joint-granular adaptation and calibration framework for domain-specific Key Information Extraction (KIE) in visually rich scanned documents. By integrating synthetic annotation pipelines with a joint-granular architecture, SynJAC substantially reduces reliance on large-scale manual annotation while preserving high performance across diverse KIE settings. Quantitative evaluations demonstrate that SynJAC achieves up to a 20\% improvement over strong baseline methods in few-shot handwritten-scanned scenarios (FormNLU-H) and maintains over 90\% accuracy even with fewer than 10\% labelled samples, confirming that synthetic structural and semantic signals can serve as a viable substitute for extensive expert annotation. Furthermore, the proposed layout embedding mechanism (L2V) yields consistent performance gains of +1–5\%, highlighting the importance of layout-aware visual cues for adaptation under noisy OCR and heterogeneous document formats. The domain adaptation strategies (SDS, SST, SIT) additionally contribute 3–8\% incremental improvement beyond the joint-grained framework alone, demonstrating the benefit of aligning fine- and coarse-grained cues during adaptation.
While SynJAC yields stable improvements across datasets, limitations persist. In particular, synthetic annotations generated via OCR parsing and LLM-based instruction tuning may introduce structural noise or semantic inconsistencies, which can occasionally affect performance in highly irregular document classes. Additionally, the current framework requires heuristic configuration, such as selecting adaptation epochs and fusion strategies, which may make it sensitive to dataset characteristics, especially under extreme noise or distribution shifts.}

\noindent \revise{\textbf{Limitations and Future Work}
Looking forward, the quantitative findings of this study open several promising research directions. First, the demonstrated scalability of synthetic annotation suggests a pathway toward fully autonomous data curation pipelines, where multimodal LLMs and agentic systems iteratively refine annotations, validation, and adaptation. Second, the results motivate theoretical investigation into noise-aware learning under synthetic supervision, particularly modelling confidence, noise propagation, and reliability of synthetic annotations across document types. Third, extending SynJAC to multilingual, multimodal, and cross-domain document ecosystems, including legal, medical, government or compliance text, may enable generalizable foundation models for VRDU. Finally, integrating SynJAC with emerging agentic LLM-based workflow automation frameworks could enable real-time document intelligence systems that self-adapt with minimal human intervention.}

\bibliographystyle{elsarticle-num} 
\bibliography{main}

\appendix
\newpage

\section{Additional Implementation Details}
\begin{table}[h]
\centering
\footnotesize
\begin{adjustbox}{max width =\linewidth}

\begin{tabular}{l|c|c|c|l}
\hline
\textbf{Fine-grained} & \textbf{Coarse-Grained} & \textbf{Configure} & \textbf{\# Para} & \textbf{\# Trainable} \\
\hline
LiLT & N/A & Baseline & 130,169,799 & 130,169,799 \\ \hline
\multirow{3}{*}{LayoutLMv3}  & N/A & Baseline & 125,332,359 & 125,332,359 \\ \cline{2-5}
&\multirow{2}{*}{LXMERT} & JG-Encoders & 393,227,514 & 19,586,415 \\
& & {\textbf{JG-$\mathcal{E}$\&$\mathcal{D}$}}  & \textbf{440,494,842} & \textbf{66,853,743} \\
\cline{3-5}
 \cline{3-5}
 \cline{2-2}
 \cline{4-5}
\hline
\end{tabular}
\end{adjustbox}
\caption{Model configurations and parameters. SynJAC is built on top of LayoutLMv3 and LXMERT following joint-grained encoder and task-specific decoders, which is bolded.}
\label{tab:num_parameters}
\end{table}

\begin{table}[h]
\centering
\footnotesize
\begin{adjustbox}{max width =\linewidth}
\label{tab:runtime_stats}
\begin{tabular}{lcccc}
\toprule
\textbf{Dataset} & \textbf{BSz} & \textbf{SDS / STS} & \textbf{FT (per epoch)} & \textbf{Inf. Time} \\
\midrule
\texttt{CORD}     & 2 & $\sim$12 min  & $\sim$35 sec $\times$ 60  & $\sim$25 sec \\
\texttt{Form-NLU} & 2 & $\sim$2 hours & $\sim$3 min $\times$ 10   & $\sim$25 sec \\
\bottomrule
\end{tabular}
\end{adjustbox}
\caption{Abbreviations: BSz = Batch Size, SDS = Synthetic Data Synthesis, STS = Synthetic Task Sampling, FT = Fine-Tuning, Inf. = Inference.}
\label{tab:time_effciency}
\end{table}

\label{app:more_llm_related}
\section{LLM Testing Details}
\subsection{LLM Testing Prompt Details}

The prompt details for each employed LLM/MLLM within the two datasets are provided in Table~\ref{tab:llm_prompts}. 
\label{app:llm_prompts}
\begin{table*}[ht]
\centering
\begin{adjustbox}{max width=\textwidth}
\begin{tabular}{>{\arraybackslash}m{1.8cm}>{\arraybackslash}m{19cm}>{\arraybackslash}m{1cm}}
\toprule
\textbf{Model} & \textbf{Prompt} & \textbf{Image} \\ \midrule

GPT-3.5 & \textit{Context: \{\} \textbackslash n Above is the context of the target form document, please extract the \{\} \textbackslash n, the output format strictly follow: Value: xxx} & N \\ \midrule
GPT-4o-t & \textit{Context: \{\} \textbackslash n Above is the context of the target form document, please extract the \{\} \textbackslash n , the output format strictly follow: Value: xxx} & N \\ \midrule
LLAVA1.5 & \textit{USER: Below image is the target form image. \textless{}image\textgreater{}\ \textbackslash n Context: \{\} \textbackslash n Above is the context of the target form document, please extract the \{\} only \textbackslash n, the output format strictly follow: \textbackslash n ASSISTANT:} & Y \\ \midrule
QWen-VL & \textit{Below image is the target form image. \textless{}image\textgreater{}\textbackslash n Context: \{\} \textbackslash n Above is the context of the target form document, please extract the \{\} only \textbackslash n, the output format should strictly follow: \textbackslash n Answer:} & Y \\ \midrule
xGen-MM & \textit{Context: \{\} \textbackslash n Above is the context of the target form document, which is \{\} \textbackslash n, output the answer only: \textbackslash n Answer: } & Y \\ \midrule
GPT-4o-v & \textit{Below image is the target form image. \textless{}image\textgreater{} Context: \{\} \textbackslash n Above is the document image and context of the target form document, please extract the \{\} \textbackslash n, the output format strictly follow: Value: xxx} & Y \\ \bottomrule
\end{tabular}
\end{adjustbox}
\caption{Comparison of prompts and image utilization across different LLMs/MLLMs.}
\label{tab:llm_prompts}
\end{table*}

\subsection{Qualitative Analysis: Limitations of LLM/MLLMs}
\label{app:case_study}

\begin{figure*}[h]
  \centering
  \includegraphics[width=\linewidth]{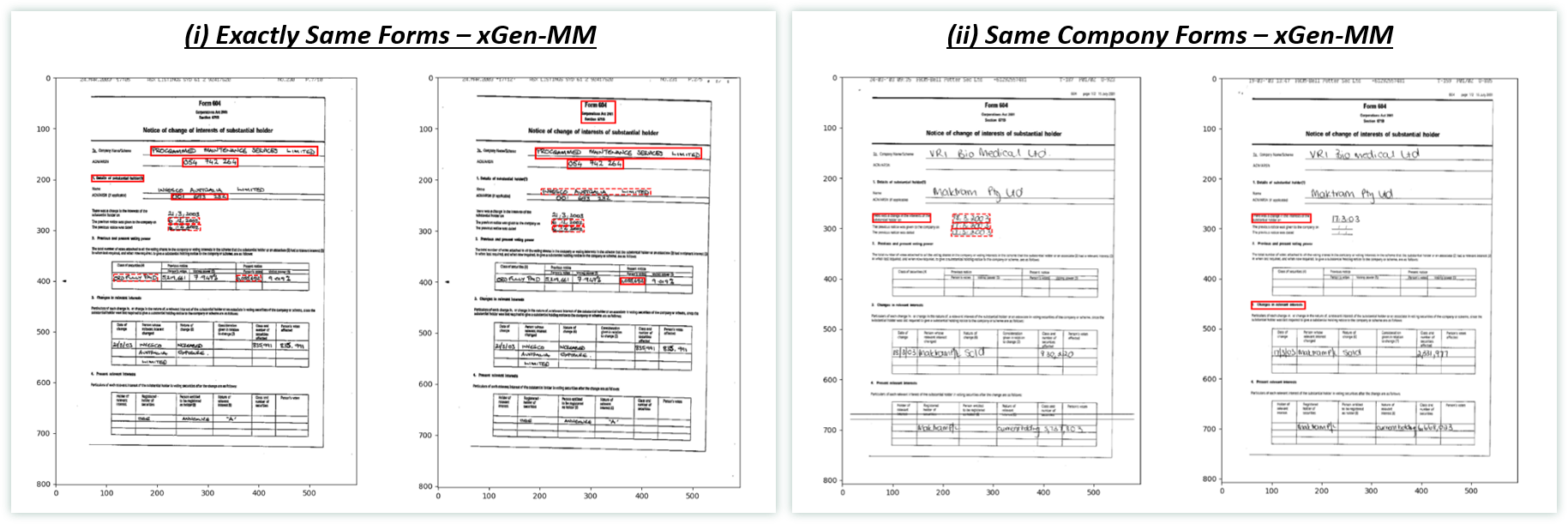}
  \caption{FormNLU sample with LLM-based document understanding (Inconsistency).}
  \label{fig:llm_inconsistency}
\end{figure*}

\begin{figure*}[h]
  \centering
  \begin{subfigure}[b]{0.48\linewidth}
    \centering
    \includegraphics[width=\linewidth]{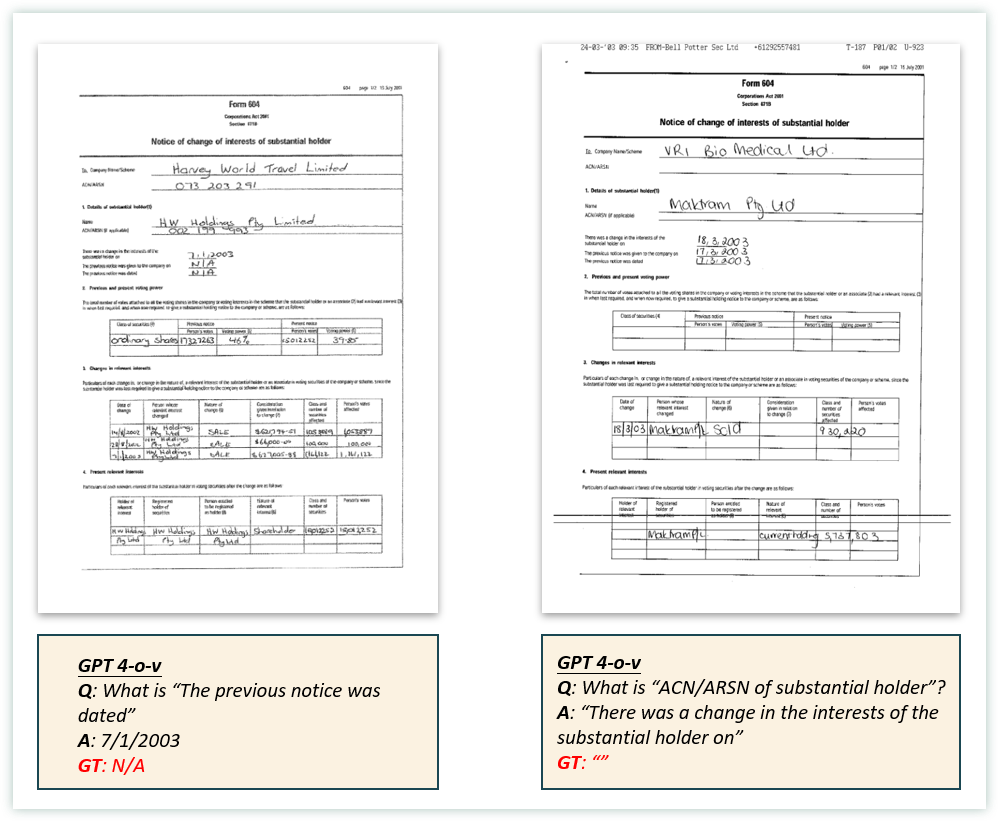}
    \caption{FormNLU sample with LLM-based document understanding (Lack of Contextual Understanding)}
    \label{fig:llm_context_understanding}
  \end{subfigure}
  \hfill
  \begin{subfigure}[b]{0.48\linewidth}
    \centering
    \includegraphics[width=\linewidth]{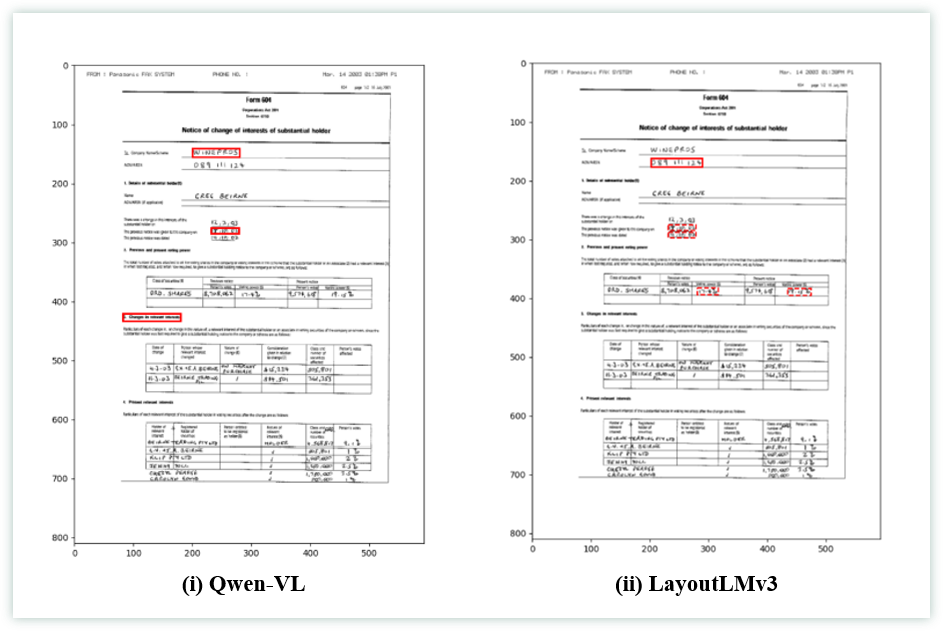}
    \caption{FormNLU LLM case study: Lack of Layout Interpretation}
    \label{fig:llm_layout_interpretation}
  \end{subfigure}
  \caption{Challenges in LLM-based document understanding in FormNLU: (a) lack of contextual understanding, (b) lack of layout interpretation.}
  \label{fig:llm_formnlu_issues}
\end{figure*}
\textbf{Layout/Structure Interpretation} LLMs excel at processing unstructured text but struggle with understanding the spatial relationships and visual structures in form-based documents. This limitation results in misaligned content, missed logical groupings, and poor performance in tasks requiring precise layout comprehension, such as interpreting complex templates or extracting values from nested structures, as shown in Figure \ref{fig:llm_layout_interpretation}.

\textbf{Inconsistency} LLMs frequently produce inconsistent outputs when handling form-based documents, generating conflicting associations for the same key-value pairs or contradicting themselves across different sections. This lack of coherence highlights their difficulty in maintaining logical consistency in structured content interpretation. For example, as shown in Figure \ref{fig:llm_inconsistency}, the LLM classifies differently between the exactly same form or the same company forms with the same person's hand writing. The same limitation was there in the receipt dataset, CORD.

\textbf{Lack of Contextual Understanding}
LLMs often generate incorrect answers by relying on superficial patterns rather than understanding contextual relationships within the document. This results in confusion between unrelated elements, making LLMs unsuitable for accurately processing structured documents that require deeper contextual and spatial alignment, as shown in Figure \ref{fig:llm_context_understanding}

\section{Supplementary of Case Studies}
 Additional supplementary materials and comprehensive analyses are provided herein for further insights.

\begin{figure*}[h]
  \centering
  \includegraphics[width=\linewidth]{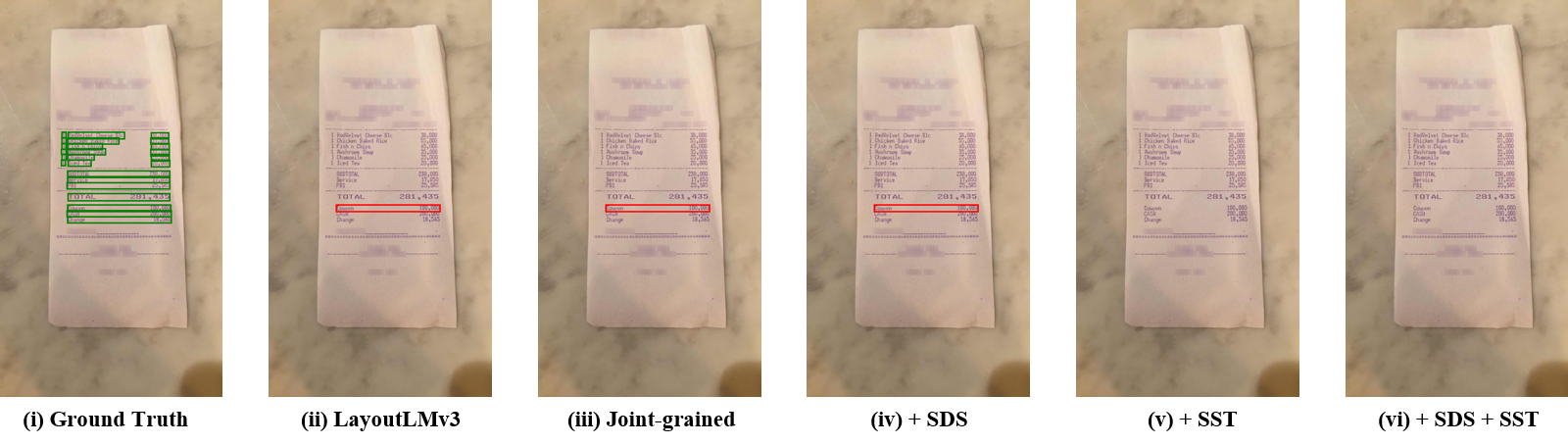}
  \caption{Real-world CORD dataset sample: (i) Ground truth key information highlighted in green. (ii) - (iv) Incorrect predictions marked with red rectangles under various configurations. (v,vi) The best performance was achieved \textbf{after applying SST} to extract all key information correctly.}

  \label{fig:case_study1}
\end{figure*}

\begin{figure*}[h]
  \centering
  \includegraphics[width=\linewidth]{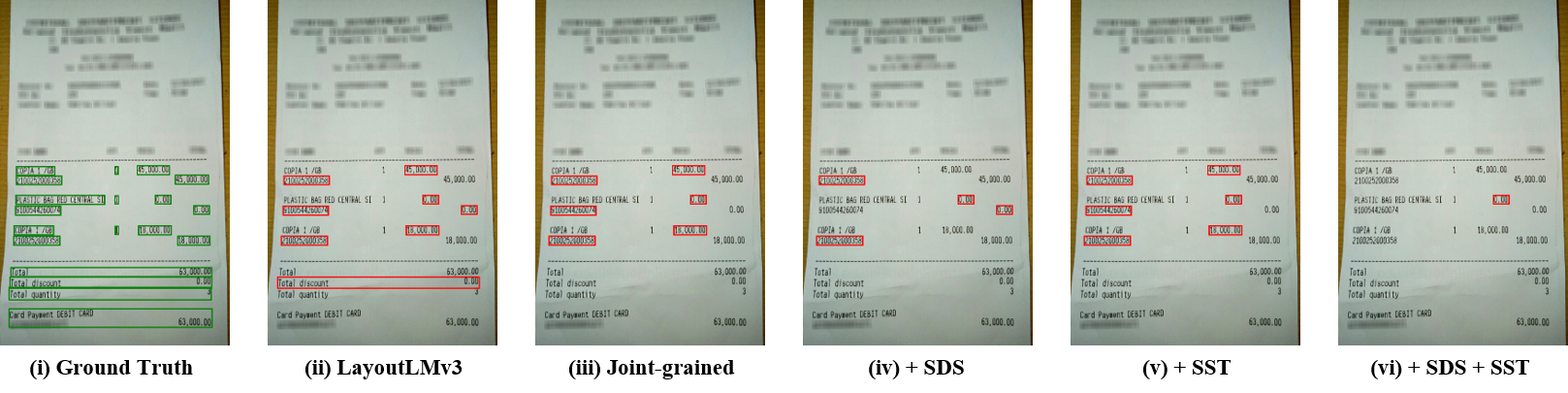}
  \caption{Real-world CORD dataset sample: (i) Ground truth key information highlighted in green. (ii) - (vi) Incorrect predictions marked with red rectangles under various configurations. (vi) The best performance was achieved using two domain adaptation methods, with only one incorrect predictions. Compared to the fine-grained-only baseline LayoutLMv3, the Joint-grained framework effectively reduces the number of incorrect cases. The \textbf{application of SDS} further decreases erroneous predictions. While the number of errors remains unchanged after applying SST, \textbf{combining SST with SDS} leads to improved robustness.}

  \label{fig:case_study2}
\end{figure*}

\begin{figure*}[h]
  \centering
  \includegraphics[width=\linewidth]{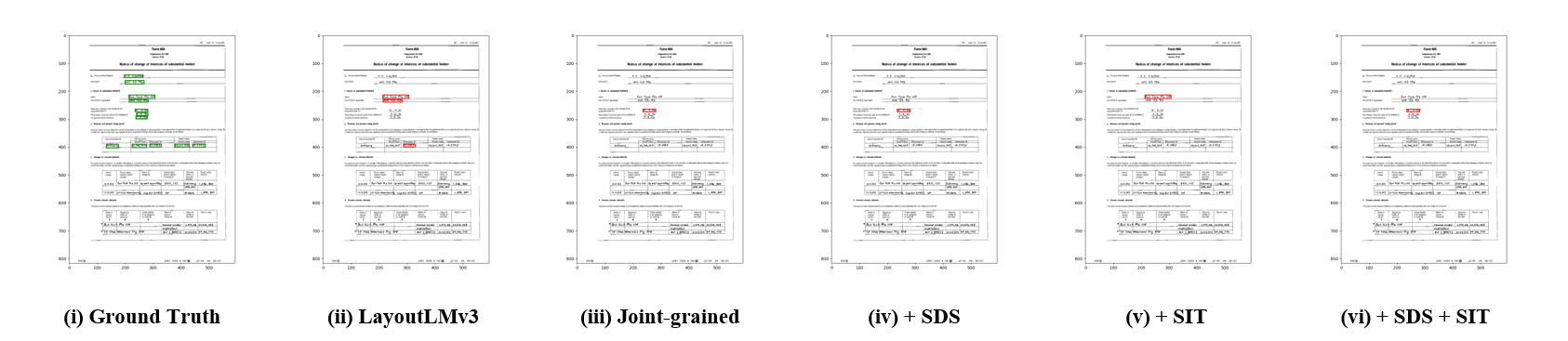}
  \caption{Real-world FormNLU handwritten dataset sample: (i) Ground truth key information highlighted in green. (ii) - (vi) Incorrect predictions marked with red rectangles under various configurations. \textbf{Joint-grained framework} could effectively reduce the number of incorrect predictions.}
  \label{fig:case_study_formnlu_h1}
\end{figure*}

\end{document}